\documentclass[lettersize,journal]{IEEEtran}
\usepackage{amsmath,amsfonts}
\usepackage{algorithmic}
\usepackage{algorithm}
\usepackage{array}
\usepackage[caption=false,font=normalsize]{subfig}
\usepackage{textcomp}
\usepackage{stfloats}
\usepackage{url}
\usepackage{verbatim}
\usepackage{graphicx}
\usepackage{cite}
\usepackage{booktabs}       
\usepackage{amsfonts}       
\usepackage{nicefrac}       
\usepackage{microtype}      
\usepackage{xcolor}         
\usepackage{array,multirow}
\usepackage{float}
\usepackage{graphicx}
\usepackage{subfig}
\usepackage{amsmath}
\usepackage{wrapfig}

\definecolor{myblue}{RGB}{158, 199, 232}
\definecolor{mygreen}{RGB}{73, 151, 122}

\begin{document}

\title{MDFlow: Unsupervised Optical Flow Learning by Reliable Mutual Knowledge Distillation}

\author{Lingtong Kong, ~~ Jie Yang
	\thanks{The authors are with the Institute of Image Processing and Pattern Recognition, Department of Automation, Shanghai Jiao Tong University, Shanghai 200240, China. (e-mail: ltkong@sjtu.edu.cn, jieyang@sjtu.edu.cn).}
}



\maketitle

\begin{abstract}
	Recent works have shown that optical flow can be learned by deep networks from unlabelled image pairs based on brightness constancy assumption and smoothness prior. Current approaches additionally impose an augmentation regularization term for continual self-supervision, which has been proved to be effective on difficult matching regions. However, this method also amplify the inevitable mismatch in unsupervised setting, blocking the learning process towards optimal solution. To break the dilemma, we propose a novel mutual distillation framework to transfer reliable knowledge back and forth between the teacher and student networks for alternate improvement. Concretely, taking estimation of off-the-shelf unsupervised approach as pseudo labels, our insight locates at defining a confidence selection mechanism to extract relative good matches, and then add diverse data augmentation for distilling adequate and reliable knowledge from teacher to student. Thanks to the decouple nature of our method, we can choose a stronger student architecture for sufficient learning. Finally, better student prediction is adopted to transfer knowledge back to the efficient teacher without additional costs in real deployment. Rather than formulating it as a supervised task, we find that introducing an extra unsupervised term for multi-target learning achieves best final results. Extensive experiments show that our approach, termed MDFlow, achieves state-of-the-art real-time accuracy and generalization ability on challenging benchmarks. Code is available at \url{https://github.com/ltkong218/MDFlow}.
\end{abstract}

\begin{IEEEkeywords}
	Optical flow, Unsupervised learning, Knowledge distillation, Real time
\end{IEEEkeywords}

\section{Introduction}
\IEEEPARstart{O}{ptical} flow estimation is a longstanding and fundamental task in computer vision, usually serving as one building block for a wide range of downstream video processing tasks, including video inpainting~\cite{9241798}, frame interpolation~\cite{Kong_2022_CVPR,Liu_2022_ICIP} and video stabilization~\cite{7508986}. Traditional approaches~\cite{10.5555/888857,5206697,5539939,8434339} usually cast optical flow estimation into an energy optimization problem, which requires time-consuming iterations and heavy computation burden, hindering them for real-time applications. Inspired by great success of deep learning on various computer vision tasks, an increasing number of works~\cite{7410673,8099662,8579029,Hur:2019:IRR,teed2020raft,NEURIPS2020_add5aebf} concentrate on making architectural progress based on convolutional neural networks (CNNs), aiming at accurate, generalization and efficiency. To achieve this goal, supervised learning methods rely on a massive amount of labeled data, which is usually synthesized by computer graphic engines~\cite{7410673,7780807,humanflow}, because of the extremely prohibitive cost to get ground truth in the wild. However, the inherent domain gap in image style and motion scene structure will damage model performance when deploying for real applications~\cite{flowdata}.

\begin{figure}[t]
	\centering
	\vspace{2mm}
	\includegraphics[width=0.9\linewidth]{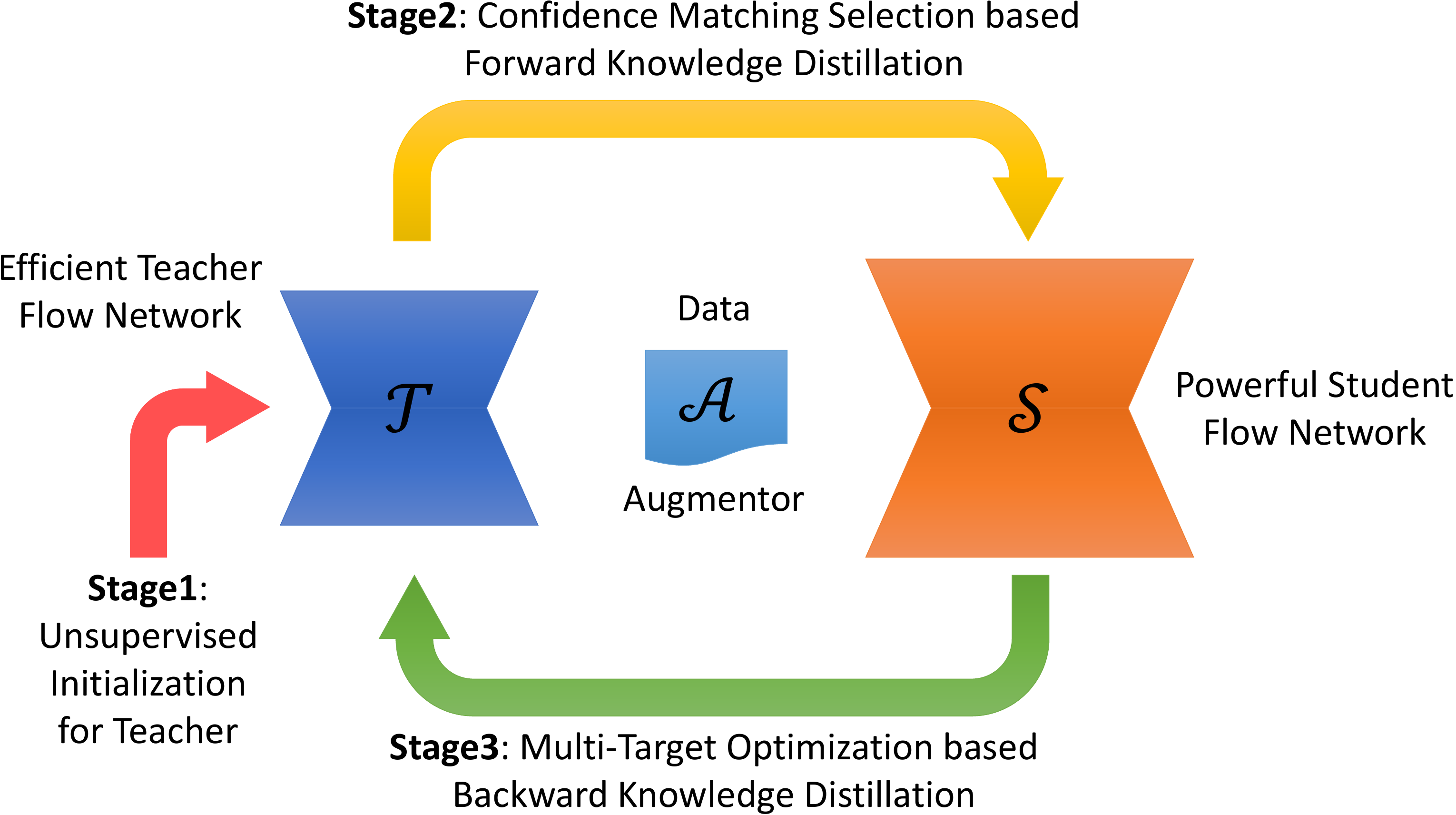}
	\vspace{-1mm}
	\caption{\textbf{Overview of proposed MDFlow algorithm for unsupervised optical flow.} Our mutual knowledge distillation approach includes an unsupervised teacher initialization stage, a confidence matching selection based reliable forward knowledge distillation stage and a multi-target optimization based reliable backward knowledge distillation stage. Each stage can improve current flow prediction accuracy while maintaining good generalization ability by employing the data augmentor $\mathcal{A}$. Thanks to the decouple nature of our framework, we can adopt any efficient teacher network $\mathcal{T}$ with any initialization method, and employ any powerful student network $\mathcal{S}$ without worrying about additional complexity in final usage. After this bidirectional mutual distillation procedure, the efficient teacher flow network is employed for real deployment.}
	\label{fig:0}
	\vspace{-1mm}
\end{figure}

One alternative pipeline is to leverage countless unlabeled video sequences for unsupervised learning, which could presumably produce satisfying results without suffering from any domain mismatch. These methods usually build objective functions based on brightness constancy and spatial smoothness~\cite{10.1007/978-3-319-49409-8_1,10.5555/3298239.3298457}, and further include occlusion handling mechanism~\cite{Meister:2018:UUL,8578611}. Most recently, the state-of-the-art ARFlow~\cite{Liu_2020_CVPR}, UFlow~\cite{10.1007/978-3-030-58536-5_33} and UPFlow~\cite{Luo_2021_CVPR} propose to integrate augmentation regularization term into each iteration step for continuous improvement. Although their self-learning approach can improve flow accuracy, the unavoidable mismatch in predicted pseudo labels can mislead the network conversely. As a result, the optimization procedure is blocked by such match noise even with more training iterations. This puts up a question: \textit{Can we decouple such mismatching regions in augmentation regularization for further accuracy improvement?}

\IEEEpubidadjcol

On the other hand, comparing with recent progress in deep flow architectures~\cite{teed2020raft,NEURIPS2020_add5aebf}, most of recent top unsupervised approaches~\cite{Liu:2019:SelFlow,Liu_2020_CVPR,10.1007/978-3-030-58586-0_11,10.1007/978-3-030-58536-5_33} adopt the PWC-Net~\cite{8579029} as flow estimation model, for its well-behaved speed-accuracy trade-off and efficiency in real deployment. Intuitively, we can achieve better accuracy by naively replacing PWC-Net~\cite{8579029} with superior flow architecture, such as RAFT~\cite{teed2020raft}, like the current state-of-the-art method SMURF~\cite{Stone_2021_CVPR}. However, this is at the cost of extra computation and delay during inference. In addition, more complex structure may lead to lower performance when guided by unsupervised losses~\cite{10.1007/978-3-030-58536-5_33}, which brings another question: \textit{Can unsupervised optical flow benefit from recent and future advanced flow architecture without introducing additional cost at inference?}

In this paper, we jointly address the above two questions with a novel unsupervised optical flow learning framework, termed MDFlow, by exploring reliable knowledge distillation between the teacher and student networks for alternative improvement. A high-level abstract of proposed mutual distillation framework is depicted in Figure~\ref{fig:0}. To answer the first question, we primarily take UFlow~\cite{10.1007/978-3-030-58536-5_33} loss functions to pretrain a teacher network as an initialization, which generates pseudo labels for image pairs. Then, a confidence mask based on residuals of census transform~\cite{10.1007/BFb0028345} is proposed to select reliable matching for training the student network with diverse data augmentation. This usually yields a student model outperforming the teacher. Concerning on the second question, we can select a stronger student to fully learn from our reliable proxy ground truth, which can in turn transfer its better knowledge back to the teacher model. Thanks to the decouple nature of our teacher student networks and mutual knowledge distillation strategy, our framework enjoys both advanced flow architecture for sufficient learning on reliable pseudo labels and efficient inference by employing a lightweight and fast teacher network. Last but not least, we will show that confidence mask plays a key role in improving student performance, and formulating knowledge distillation back to the teacher as a multi-target optimization objective yields the best final results.

Our contributions are summarized as: 
\begin{itemize}
	\item
	We present a novel mutual knowledge distillation framework on unsupervised optical flow for improved performance without additional cost during inference.
	\item
	Newly proposed reliable matching selection mechanism and multi-target joint learning pipeline are proved to be effective in forward and backward flow knowledge distillation processes.
	\item
	Our approach achieves state-of-the-art accuracy and generalization ability on Sintel~\cite{Butler:ECCV:2012} and KITTI 2015~\cite{Menze2015CVPR} benchmarks compared with other pyramid-based methods.
\end{itemize}

\section{Related Work}
\subsection{Deep Optical Flow Architecture}
Finding dense correspondences between a pair of time adjacent frames, namely optical flow estimation, has been studied for decades for its fundamental role in many downstream vision tasks. FlowNet~\cite{7410673} and its successor FlowNet2~\cite{8099662} are the first attempt to apply deep learning methods for optical flow estimation, which directly regress flow field based on the encoder-decoder U-shape network or its cascaded version. Inspired by traditional coarse-to-fine pipeline~\cite{5206697,7780984,8434339}, SPyNet~\cite{8099774}, PWC-Net~\cite{8579029} and LiteFlowNet~\cite{8579034} employ pyramid, warping and cost volume into end-to-end learning and achieve impressive real-time performance. After that, IRR-PWC~\cite{Hur:2019:IRR} iteratively and jointly estimates residual flow and occlusion with shared estimators across pyramid levels. The work~\cite{8846749} introduces a dual self-attention module to improve original pyramid flow network. MaskFlowNet~\cite{zhao2020maskflownet} and OAS-Net~\cite{9413531} explore to resolve ambiguous matching caused by warping operation. As for more efficient architecture, FDFlowNet~\cite{9191101} introduces U-shape backbone and partial fully connected decoder for compact structure. FastFlowNet~\cite{Kong_2021_ICRA} constructs center dense dilated correlation layer and shuffle block decoder to reduce computation and first achieves real-time performance on embedded systems. Recently, DICL-Flow~\cite{NEURIPS2020_add5aebf} transfers concatenated feature volume in stereo matching to optical flow and propose displacement-invariant cost learning. Different from above coarse-to-fine pipeline, RAFT~\cite{teed2020raft} first calculates all pairs' correspondences and then introduces a recurrent module to estimate residual flow and update multi-scale matching cost simultaneously, which achieves significant accuracy improvement. Later on, GMA~\cite{Jiang_2021_ICCV} introduces a global motion aggregation module based on transformer to improve recurrent-based flow architecture and obtains state-of-the-art accuracy.

\subsection{Learning Unsupervised Optical Flow}
Due to prohibitive cost to acquire optical flow ground truth of real-world images, alternative approaches try to leverage countless unlabeled video sequences for unsupervised learning. These methods build objective functions based on brightness constancy assumption and local smoothness prior~\cite{10.1007/978-3-319-49409-8_1,10.5555/3298239.3298457}. Then, UnFlow~\cite{Meister:2018:UUL} and OccAwareFlow~\cite{8578611} boost unsupervised performance by excluding residual calculation in reasoned occluded regions. However, the occluded areas are only guided by rigid smoothness constraint, that can damage overall accuracy. To handle this problem, DDFlow~\cite{Liu:2019:DDFlow} and SelFlow~\cite{Liu:2019:SelFlow} distill flow estimation from the teacher model to the student with random crop and occlusion hallucination in a data-driven manner, that further improves unsupervised accuracy. However, the teacher can not benefit from the improved student. STFlow~\cite{9201360} integrates variational refinement with unsuperivsed learning and propose a self-taught framework for continual improvement. SimFlow~\cite{10.1007/978-3-030-58586-0_11} explores learnable feature similarity for regulating previous census reconstruction loss. ARFlow~\cite{Liu_2020_CVPR} and UFlow~\cite{10.1007/978-3-030-58536-5_33} propose to integrate augmentation regularization into each iteration step for continuous improvement of a single model. CoT-AMFlow~\cite{Wang_CoRL_2020} develops an adaptive modulation network and adopts a co-teaching strategy for better accuracy. Later on, UPFlow~\cite{Luo_2021_CVPR} improves the upsampling unit of PWC-Net and proposes a better pyramid distillation loss. DistillFlow~\cite{9444870} trains multiple teacher models and introduces a confidence based two-stage distillation approach for improvement. OIFlow~\cite{9477059} puts up an occlusion-inpainting framework to make full use of occlusion regions. Recently, ASFlow~\cite{9625946} presents content-aware pooling and adaptive flow upsampling modules to improve pyramid-based unsupervised flow deep structure. MRDFlow~\cite{9648363} further introduces 4D correlation layer and recurrent decoder of RAFT to strengthen flow estimation network. At the same time, SMURF~\cite{Stone_2021_CVPR} replaces PWC-Net with a more powerful RAFT backbone and proposes a sequence-aware self-supervision loss to achieve state-of-the-art accruacy. However, it incurs huge computation cost and can not satisfy many real-time applications. Different from above methods, we separate and recombine multiple objective parts into a mutual distillation framework for reliable performance improvement without increasing model size and inference delay.

Another line of unsupervised optical flow methods resort to additional information beyond two input frames. Based on the setting of stereo video systems, UnOS~\cite{Wang_2019_CVPR} enforces geometry constraints among stereo depth, camera ego-motion and optical flow for mutual promotion. Flow2Stereo~\cite{Flow2Stereo} introduces data distillation into the joint learning framework of optical flow and stereo matching. Most recently, the work~\cite{Chi_2021_CVPR} shows that feature-level collaboration of the networks for optical flow, stereo depth and camera motion can outperform previous methods that only consider loss-level joint optimization. To facilitate unsupervised optical flow in challenging scenes, such as fog, rain and night, GyroFlow~\cite{Li_2021_ICCV} converts gyroscope readings into gyro field and fuse it with flow information for recovering more motion details. Different from these methods that resort to additional information, proposed MDFlow focuses on improving the basic setting by reducing matching noise and exploring better flow architecture during the decoupled bidirectional distillation procedure.

\subsection{Knowledge Distillation and Mutual Distillation}
Knowledge Distillation (KD) is usually exploited to train a compact student network by mimicking the output distribution of a pre-trained complex teacher model as well as one-hot ground-truth labels~\cite{44873}. Following variants~\cite{DBLP:journals/corr/RomeroBKCGB14,Zagoruyko2017AT,8953814,tian2019crd} mainly focus on utilizing intermediate information of teacher, such as feature maps or attention masks as extra hints for improvement. Different from above one-way transfer between a teacher and a student in knowledge distillation, Deep Mutual Learning (DML)~\cite{Zhang_2018_CVPR} finds that mutual learning of a collection of simple student networks can outperforms distillation from a more powerful yet static teacher. The work EKD~\cite{9461003} introduces an evolutionary teacher that can enable more efficient knowledge transfer by minimizing the capability gap between teacher and student. Dense Cross-layer Mutual-distillation (DCM)~\cite{10.1007/978-3-030-58555-6_18} improves the two-way Knowledge Transfer (KT) scheme by adding auxiliary classifiers to hidden layers of both teacher and student networks and building dense bidirectional KD between these classifiers. The work~\cite{NEURIPS2021_c203d8a1} also adopts mutual distillation method to encourage information exchange between the local branch and the global branch of a single network for discovering new categories. There are also some research about applying mutual learning to computer vision tasks, such as object detection~\cite{8954274,9577564} and video-based sign language recognition~\cite{9710140}. Existing mutual distillation approaches have achieved success in high-level classification tasks. However, whether mutual distillation can help the middle-level dense matching tasks has not been fully explored.

As for optical flow estimation, DDFlow~\cite{Liu:2019:DDFlow} and SelFlow~\cite{Liu:2019:SelFlow} distill flow estimation from teacher to student with several occlusion hallucination methods, which has been shown to be effective when inferring on occluded regions. DistillFlow~\cite{9444870} further improves above two-stage data distillation by introducing multiple teacher models and confidence mechanism. As for a related medical image registration task, the work CRD~\cite{9782430} distills knowledge from a feature-shifted teacher model with high resolution input to a student model with low resolution input for more efficiency. Different from their works, our MDFlow transfers knowledge between teacher and student networks mutually, so as to decouple matching outliers in augmentation regularization and exploit recent advanced architecture as student for better final prediction, while maintaining real-time inference.

\section{Method}
In this section, we first illustrate the overall pipeline of MDFlow algorithm for unsupervised optical flow, and then provide notations and unsupervised initialization approach used in our framework. Further, the reliable forward knowledge distillation process is demonstrated, where a novel confidence matching selection mechanism plays a key role. At last, we present the reliable backward knowledge distillation process for improving the accuracy of final deployed teacher model.

\begin{figure*}[t]
	\centering
	\includegraphics[width=0.98\linewidth]{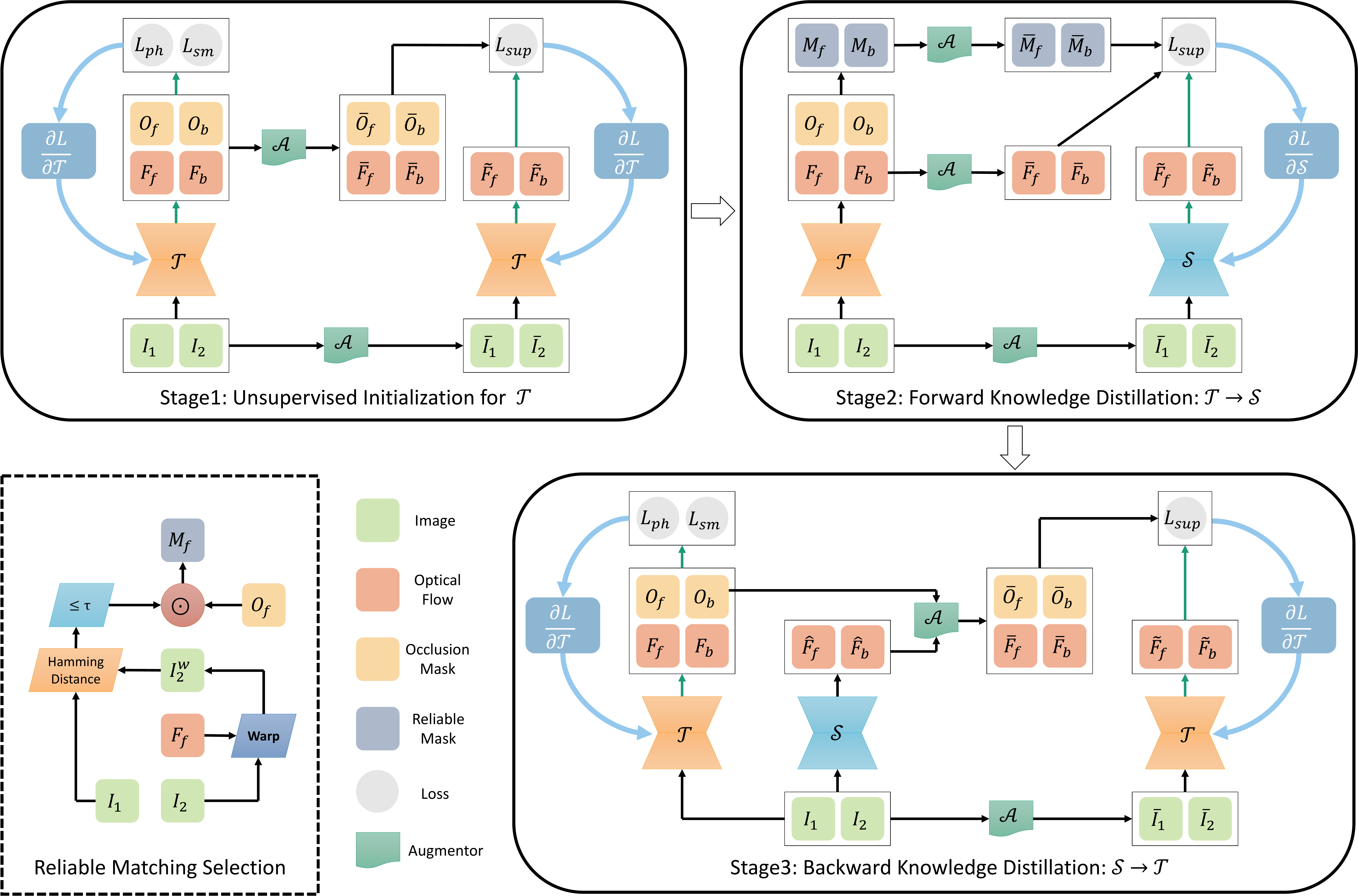}
	\caption{\textbf{Detailed framework of reliable mutual knowledge distillation for unsupervised optical flow.} \textcolor{mygreen}{Green line} denotes the forward path and \textcolor{myblue}{blue line} denotes the backward path. Networks with the same color share weights in each stage. Reliable masks $M_f, M_b$ generated in stage 2 are shown in the left bottom corner. Note that the inputs and outputs of the data augmentor $\mathcal{A}$ are detached from the gradient calculation diagram for stable learning.}
	\label{fig:1}
\end{figure*}

\subsection{Overview}
As depicted in Figure~\ref{fig:0}, our proposed MDFlow algorithm trains a teacher model $\mathcal{T}$ and a student model $\mathcal{S}$ interactively for progressive improvement in a three-stage manner. In the first stage, our teacher network can be trained by any unsupervised approach for generating noisy pseudo labels, which will be used later. Specifically, we adopt PWC-Net~\cite{8579029} as teacher and use UFlow~\cite{10.1007/978-3-030-58536-5_33} loss functions for unsupervised initialization. In the second stage, we introduce a novel reliable matching selection mechanism, that can partly remove outliers in generated pseudo labels, which will then be used to train a student network with diverse data augmentation approaches, constituting our forward knowledge distillation process. Note that we can select a stronger student architecture, such as RAFT~\cite{teed2020raft}, for sufficient learning thanks to the decouple nature in mutual knowledge distillation. In the last stage, better pseudo labels predicted by the student will be adopted to supervise the learning of original teacher model enriched by diverse data augmentation. Moreover, we can improve the final teacher performance by adding a weak unsupervised learning objective, formulating it into a multi-target manner. More details are shown in Figure~\ref{fig:1}.

\subsection{Unsupervised Initialization for Teacher Model}
Given two consecutive frames $I_1, I_2$ from unlabeled image sequences $\mathcal{I}$, unsupervised optical flow methods aim to train a flow network $\mathcal{N}(.)$ based on brightness constancy assumption and spatial smoothness prior. The first goal is usually realized as a photometric reconstruction loss:
\begin{equation}
	\begin{split}\mathcal{L}_{ph}(\mathcal{N}) = & \frac{\sum_{p}{\rho(I_1-I_2^w)\odot (1-O_f)}}{\sum_{p}{(1-O_f)}}+ \\
		& \frac{\sum_{p}{\rho(I_2-I_1^w)\odot (1-O_b)}}{\sum_{p}{(1-O_b)}},
	\end{split}
	\label{eq:1}
\end{equation}
where $\odot$ represents element-wise multiplication, $O_f, O_b$ are occlusion masks inferred by forward-backward consistency check~\cite{10.1007/978-3-642-15549-9_32,Meister:2018:UUL}, and $\rho(x)=(|x|+\epsilon)^{q}$ is a robust function with $\epsilon=0.01, q=0.4$ in all our experiments. Denoting $F_f = \mathcal{N}(I_1, I_2), F_b = \mathcal{N}(I_2, I_1)$ as predicted forward and backward flow fields, and $I_2^w = \textit{w}(I_2, F_f), I_1^w = \textit{w}(I_1, F_b)$ are flow warped reconstruction images. Following previous research~\cite{Liu:2019:DDFlow,10.1007/978-3-030-58536-5_33}, we use the soft Hamming distance of census transformed~\cite{10.1007/BFb0028345} image patches to calculate reconstruction for its robust to illuminance variation.

\begin{figure*}[t]
	\footnotesize
	\centering
	\resizebox{1.0\textwidth}{!}{
		\begin{tabular}{@{}c @{\hskip 0.02in} c @{\hskip 0.02in} c @{\hskip 0.02in} c@{}}
			\includegraphics[width=0.195\linewidth]{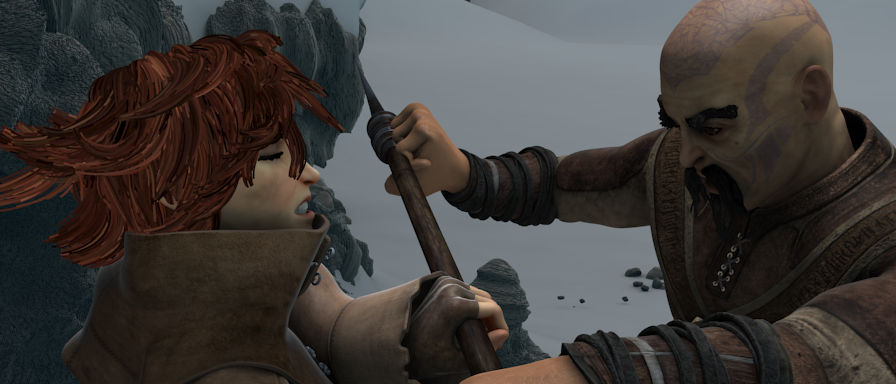}
			&
			\includegraphics[width=0.195\linewidth]{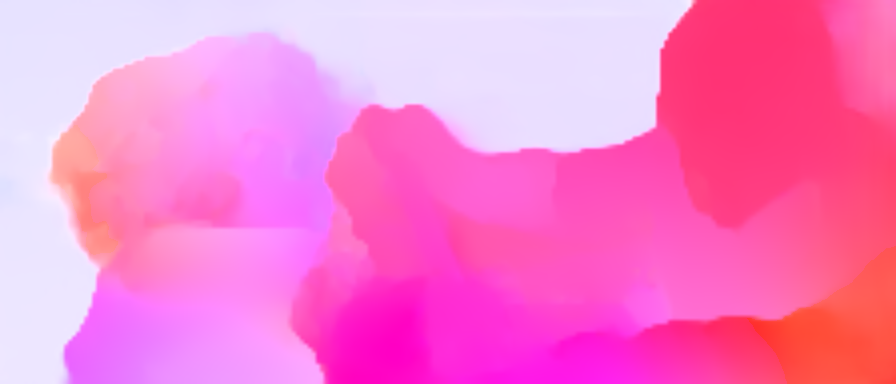}
			&
			\includegraphics[width=0.195\linewidth]{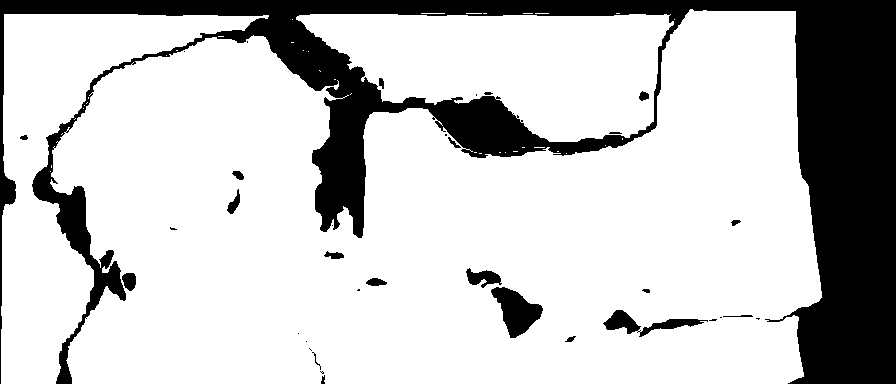}
			\vspace{-1.4mm}
			\\
			\tiny{(a) First Input Image} & \tiny{(b) Flow Prediction of UFlow~\cite{10.1007/978-3-030-58536-5_33}} & \tiny{(c) Forward Non-Occluded Mask}
			\\
			\includegraphics[width=0.195\linewidth]{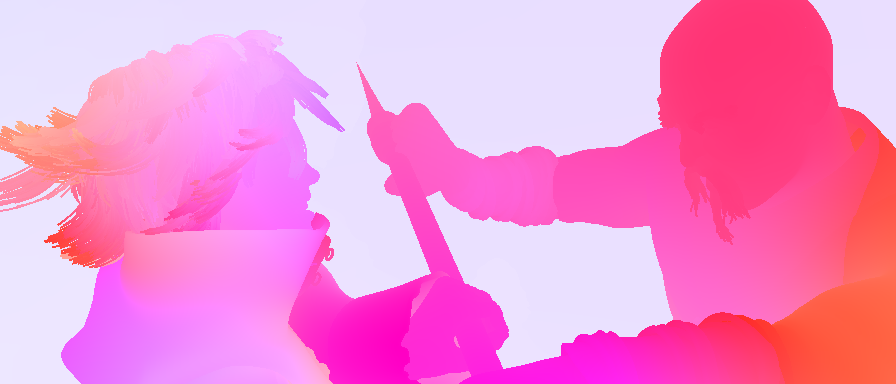}
			&
			\includegraphics[width=0.195\linewidth]{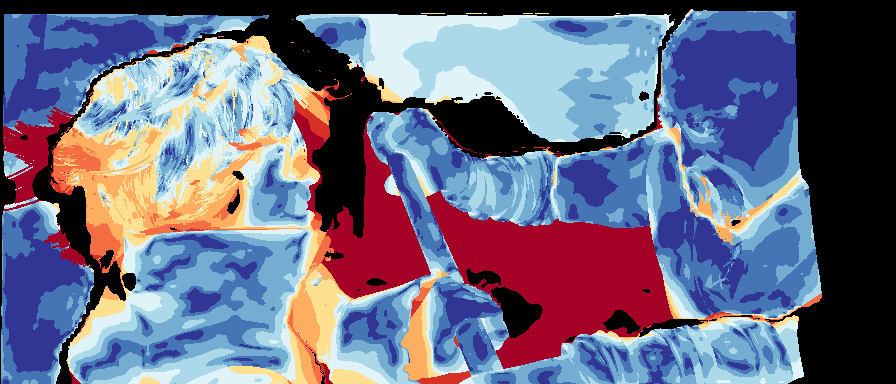}
			&
			\includegraphics[width=0.195\linewidth]{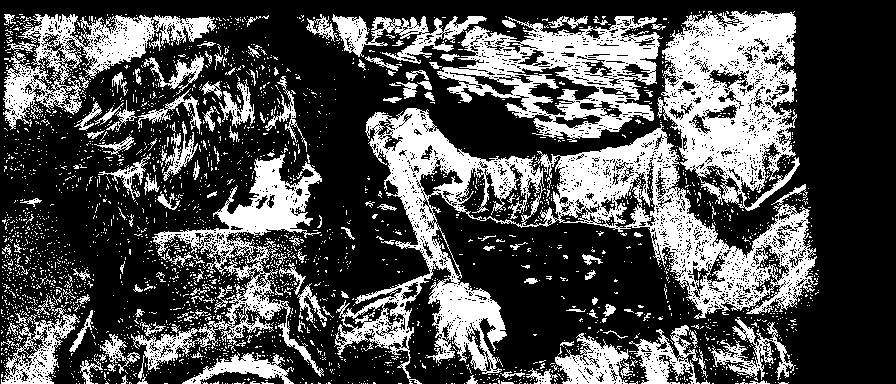}
			\vspace{-1.4mm}
			\\
			\tiny{(d) Ground Truth Flow} & \tiny{(e) Flow Error of UFlow~\cite{10.1007/978-3-030-58536-5_33}} & \tiny{(f) Proposed Reliable Matching Mask}
	\end{tabular}}
	\caption{\textbf{Problem of current superior approach and proposed reliable matching mask.} Forward non-occluded mask in (c) is calculated by forward backward consistency check. For flow prediction error visualization in (e), deeper red means larger error and deeper blue means smaller error. Best viewed in color.}
	\label{fig:2}
\end{figure*}

Due to the well-known aperture problem, above solution can be ambiguous on textureless or repetitive pattern regions, so we further introduce a loss item to constrain the spatial smoothness of predicted flow fields, which is usually reweighted by local image gradients:
\begin{equation}
	\mathcal{L}_{sm}(\mathcal{N}) = \sum_{d \in x, y} \sum_{p} (| \nabla_{d}^k F_f | e^{- \lambda | \nabla_{d}^1 I_1 |} + | \nabla_{d}^k F_b | e^{- \lambda | \nabla_{d}^1 I_2 |}),
	\label{eq:2}
\end{equation}
where $\lambda$ controls edge-aware weighting strength conditioned on reference image. $\nabla_{d}^k, k=1, 2$ stands for the $k$-th order gradient operator. In our experiments, we set $\lambda = 150$. For smoothness, following the experience in UFlow~\cite{10.1007/978-3-030-58536-5_33}, we use first order operator on Flying Chairs and Sintel, while adopting second order operator on KITTI for their better results. The reason is that there are more parallel motion about image plane in Chairs and Sintel, while more vertical motion in regard to image plane in driving scenes of KITTI. Flow fields projected from the parallel and vertical motion about image plane satisfy the first and second order smoothness respectively.

Inspired by recent work of ARFlow~\cite{Liu_2020_CVPR} and UFlow~\cite{10.1007/978-3-030-58536-5_33}, we also add a self-supervised loss for continual improvement. Denoting $\mathcal{A}$ as the data augmentor, which can jointly augment input images $I_1, I_2$, pseudo labels $F_f, F_b$ and valid masks $M_f, M_b$ by applying spatial and color transformations. Indicating $\overline{I_1}, \overline{I_2}, \overline{F_f}, \overline{F_b}$ and $\overline{M_f}, \overline{M_b}$ as corresponding augmented images, pseudo labels and valid masks, the surrogate supervision loss can be formulated as:
\begin{equation}
	\begin{split}
		\mathcal{L}_{sup}(\mathcal{N}_1 | \mathcal{N}_2, \mathcal{A})= & \frac{\sum_{p}{|\mathcal{N}_1(\overline{I_1}, \overline{I_2}) - \overline{F_f}| \odot \overline{M_f}}}{\sum_{p}{\overline{M_f}}}+ \\
		& \frac{\sum_{p}{|\mathcal{N}_1(\overline{I_2}, \overline{I_1}) - \overline{F_b}| \odot \overline{M_b}}}{\sum_{p}{\overline{M_b}}}.
	\end{split}
	\label{eq:3}
\end{equation}
Note that there are two models $\mathcal{N}_1$ and $\mathcal{N}_2$ in Eq.~\ref{eq:3}, where $\mathcal{N}_1$ is optimized conditioned on the pseudo labels $F_f, F_b$ predicted by $\mathcal{N}_2$ and augmentor $\mathcal{A}$. Following previous experience, augmented training samples are detached from gradient calculation diagram for stable learning. We adopt above loss notation for clarity of following expression. In regard to the first stage of MDFlow, we employ the off-the-shelf UFlow~\cite{10.1007/978-3-030-58536-5_33} method to initialize the teacher $\mathcal{T}$ with following objective:
\begin{equation}
	\mathcal{L}_{s1} = \mathcal{L}_{ph}(\mathcal{T}) + \lambda_1 \mathcal{L}_{sm} (\mathcal{T})+ \lambda_2 \mathcal{L}_{sup}(\mathcal{T} | \mathcal{T}, \mathcal{A}),
	\label{eq:4}
\end{equation}
where $\lambda_1, \lambda_2$ are trade-off coefficients, while $\mathcal{L}_{sup}(\mathcal{T} | \mathcal{T}, \mathcal{A})$ means that the second forward prediction is supervised by the augmented first prediction of the same model $\mathcal{T}$.

\begin{figure}[t]
	\centering
	\includegraphics[width=0.8\linewidth]{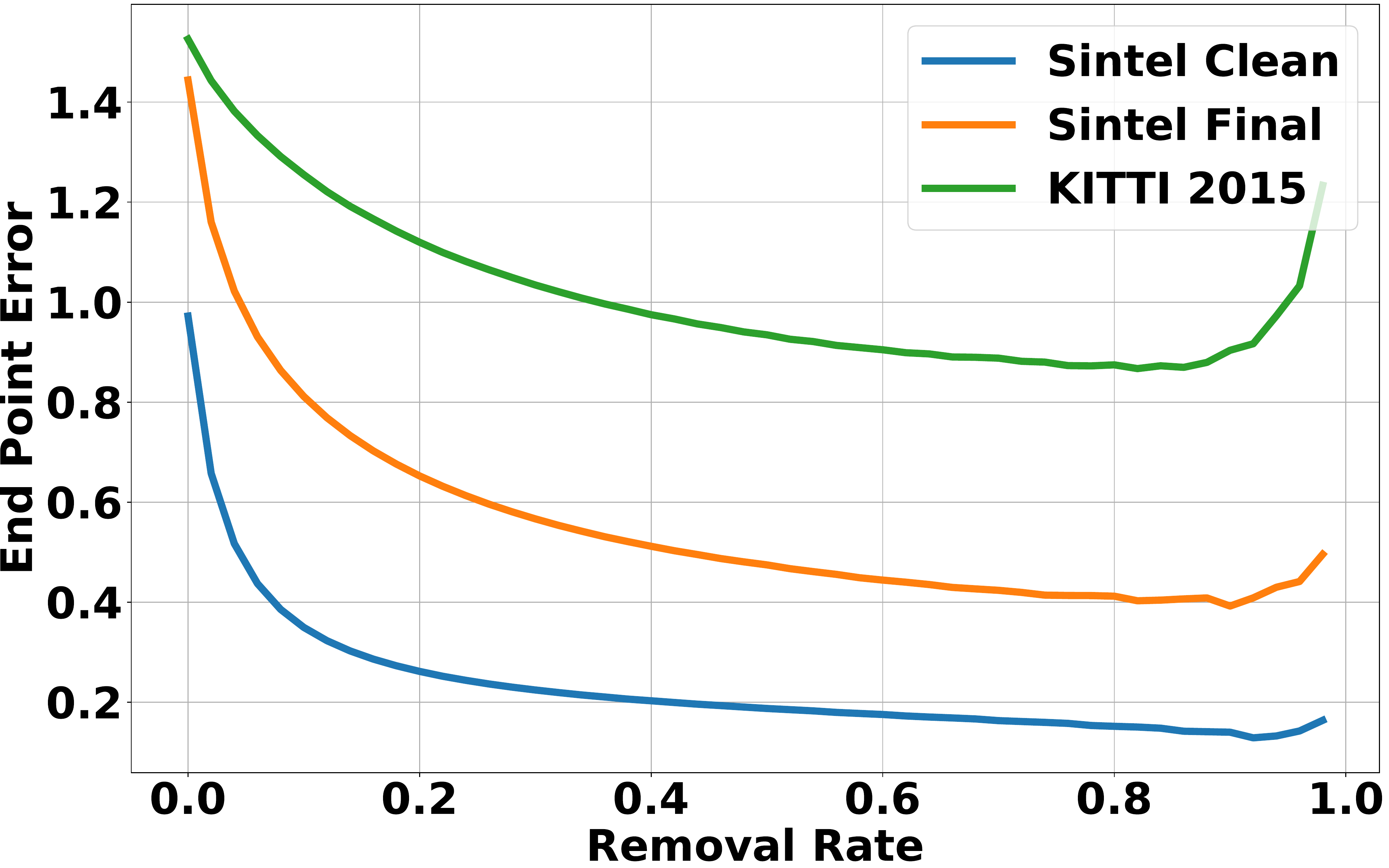}
	\caption{\textbf{Sparsification Curves on Sintel and KITTI datasets.}}
	\label{fig:3}
\end{figure}

\subsection{Reliable Forward Knowledge Distillation}
Due to shadow, exposure, textureless and repetitive patterns existed in natural images, optimal solution of photometric reconstruction will inevitably contain mismatching areas. On the other hand, state-of-the-art unsupervised optical flow approaches~\cite{Liu_2020_CVPR,10.1007/978-3-030-58536-5_33} augment predicted flow fields for a second supervision to improve performance. Concretely, they select non-occluded regions as the valid mask when calculating above self-supervised loss. As shown in Figure~\ref{fig:2}, large error regions in predicted pseudo labels, for example, the background surrounded by moving persons, are not identified by occlusion map. Thus, the augmented surrogate supervision will hinder the network from learning true displacement. To alleviate this problem, we first propose a novel confidence matching selection mechanism to find relatively reliable prediction, and then separate $\mathcal{L}_{sup}$ in Eq.~\ref{eq:4} with $\mathcal{L}_{ph}, \mathcal{L}_{sm}$ into a supervised knowledge distillation process. Holding the opinion that better matched image patches should have lower residuals, we explore to set a threshold to keep only a portion of reliable pseudo labels during forward knowledge distillation. In this work, we employ the soft Hamming distance of census transformed image patches for confidence measurement, and also include a robust function $\rho$, which does not change confidence ranking. Denote $M_f$ as forward valid mask and $\tau$ as a threshold, the selection approach can be formulated as:
\begin{equation}
	M_f(p)=\left\{
	\begin{aligned}
		1 & , & \rho(I_1(p) - I_2^w(p)) \le \tau, \\
		0 & , & \rho(I_1(p) - I_2^w(p)) > \tau.
	\end{aligned}
	\right.
	\label{eq:5}
\end{equation}

\begin{table}[t]
	\renewcommand\arraystretch{1.1}
	\caption{\textbf{Valid mask selection threshold and endpoint error on MPI Sintel and KITTI 2015 training datasets with specific removal rates.} Models are trained with loss function in Eq.~\ref{eq:4} on Sintel and KITTI training datasets respectively.}
	\label{tab:1}
	\centering
	\renewcommand{\arraystretch}{1.1}
	\tabcolsep=3.6mm
	\begin{tabular}{c|cc|cc}
		\toprule
		\multirow{2}{*}{Removal Rate} & \multicolumn{2}{c|}{MPI Sintel} & \multicolumn{2}{c}{KITTI 2015} \\
		& \multicolumn{1}{c}{Threshold} & \multicolumn{1}{c|}{EPE} & \multicolumn{1}{c}{Threshold} & \multicolumn{1}{c}{EPE} \\
		\midrule
		\multicolumn{1}{c|}{0\%} & 4.66 & 1.21 & 4.66 & 1.53 \\
		\multicolumn{1}{c|}{10\%} & 3.22 & 0.58 & 3.59 & 1.26 \\
		\multicolumn{1}{c|}{20\%} & 2.82 & 0.46 & 3.26 & 1.12 \\
		\multicolumn{1}{c|}{30\%} & 2.54 & 0.40 & 2.99 & 1.03 \\
		\bottomrule
	\end{tabular}
\end{table}

In order to set meaningful threshold for outlier removal, we analyse the variation curve of average endpoint error with removal rate on Sintel and KITTI training sets, which is controlled by $\tau$. Similar to uncertainty estimation~\cite{10.1007/978-3-030-01234-2_40}, the sparsification curves are depicted in Figure~\ref{fig:3}, while several typical removal rates with corresponding thresholds and errors are listed in Table~\ref{tab:1}. As can be seen, average endpoint error of remaining pseudo labels decreases as we gradually remove more prediction from original non-occluded mask, until removal rate reaches about $90\%$. On the other hand, larger percentage of removal means less pseudo labels for training, that can damage performance. Therefore, setting reasonable removal rate for better trade-off plays a key role in forward distillation process. We leave this question in experiment part for detailed discussion. Guided by the above confidence matching selection mask $M_f, M_b$, reliable forward knowledge distillation can be formulated as:
\begin{equation}
	\mathcal{L}_{s2} = \mathcal{L}_{sup}(\mathcal{S} | \mathcal{T}, \mathcal{A}).
	\label{eq:6}
\end{equation}

\subsection{Reliable Backward Knowledge Distillation}
Due to the decouple characteristic of proposed mutual distillation framework, we can employ a stronger flow architecture as student $\mathcal{S}$ than the original teacher $\mathcal{T}$, so as to learn from reliable pseudo labels sufficiently. This process is different from traditional knowledge distillation~\cite{44873,aleotti2020learning} where a lightweight student usually distills knowledge from a large teacher model. Therefore, our approach further includes a backward distillation stage, aiming to transfer better student knowledge back to the efficient teacher model. As a result, proposed MDFlow does not add extra computation cost and inference delay in real deployment, compared with other unsupervised approaches.

To achieve this goal, one may follow the supervision loss of Eq.~\ref{eq:6} used in stage 2, by exchanging $\mathcal{T}$ and $\mathcal{S}$. However, due to the limited learning ability of efficient optical flow network, and diverse strong augmentation imposed on pseudo labels, training data for the efficient teacher $\mathcal{T}$ is blended with augmentation noise, which deviates from the distribution of real-world dynamic scenes. On the other hand, we can not remove the augmentor $\mathcal{A}$ during backward distillation, since data augmentation has been proved to be one crucial step for excellent performance~\cite{8621052,Bar-Haim_2020_CVPR}. To deal with this problem, we establish a multi-target learning manner for reliable backward knowledge distillation, integrating specific domain knowledge on optical flow task. Specifically, unsupervised photometric and smoothness losses based on original scene structure are introduced as regularization. In short, reliable backward distillation of stage 3 can be written as:
\begin{equation}
	\mathcal{L}_{s3} = \mathcal{L}_{sup}(\mathcal{T} | \mathcal{S}, \mathcal{A}) + \lambda_3 \mathcal{L}_{ph}(\mathcal{T}) + \lambda_4 \mathcal{L}_{sm}(\mathcal{T}),
	\label{eq:7}
\end{equation}
where $\lambda_3, \lambda_4$ are weight coefficients controlling unsupervised regularization intensity, and non-occluded masks $1-O_f, 1-O_b$ of $\mathcal{T}$ are adopted as valid masks in $\mathcal{L}_{sup}$.

\begin{figure*}[t]
	\footnotesize
	\centering
	\resizebox{1.0\textwidth}{!}{
		\begin{tabular}{@{}c @{\hskip 0.01in} c @{\hskip 0.01in} c @{\hskip 0.01in} c @{\hskip 0.01in} c@{}}
			\includegraphics[width=0.195\linewidth]{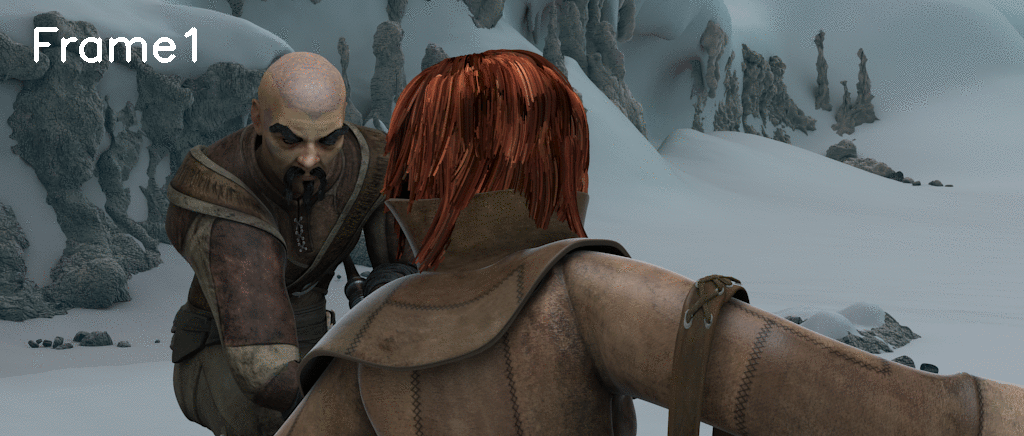}
			&
			\includegraphics[width=0.195\linewidth]{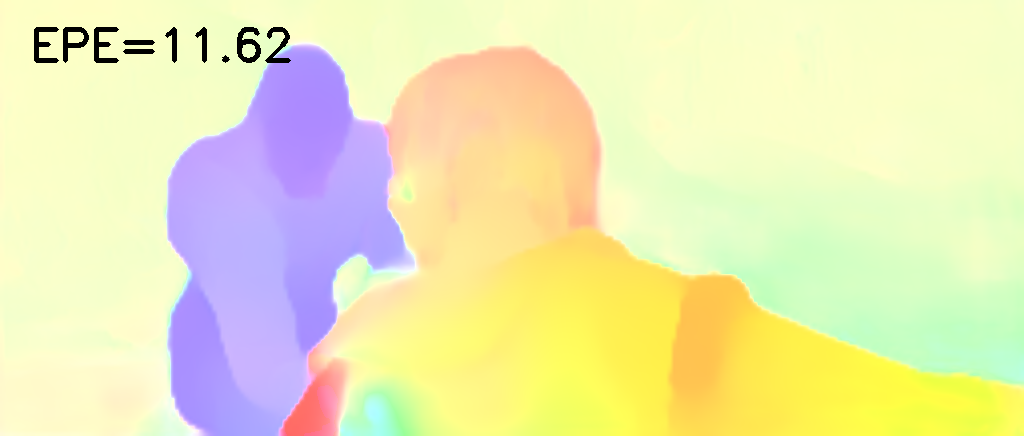}
			&
			\includegraphics[width=0.195\linewidth]{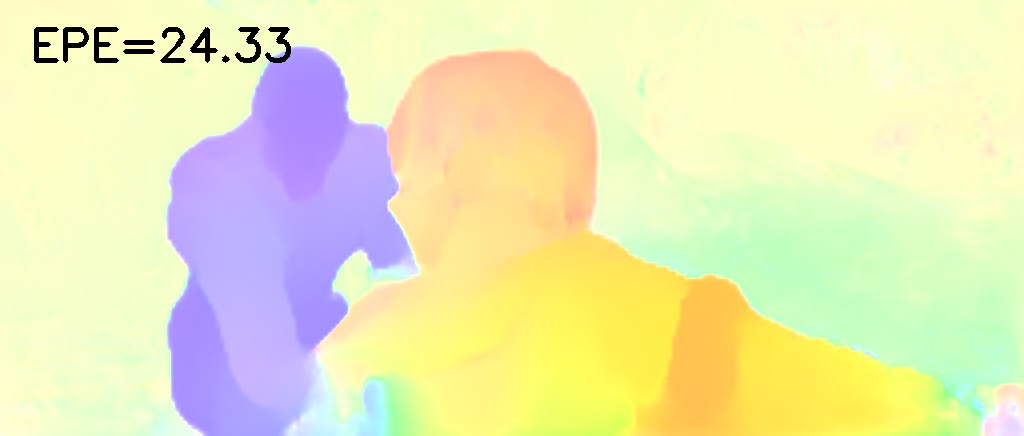}
			&
			\includegraphics[width=0.195\linewidth]{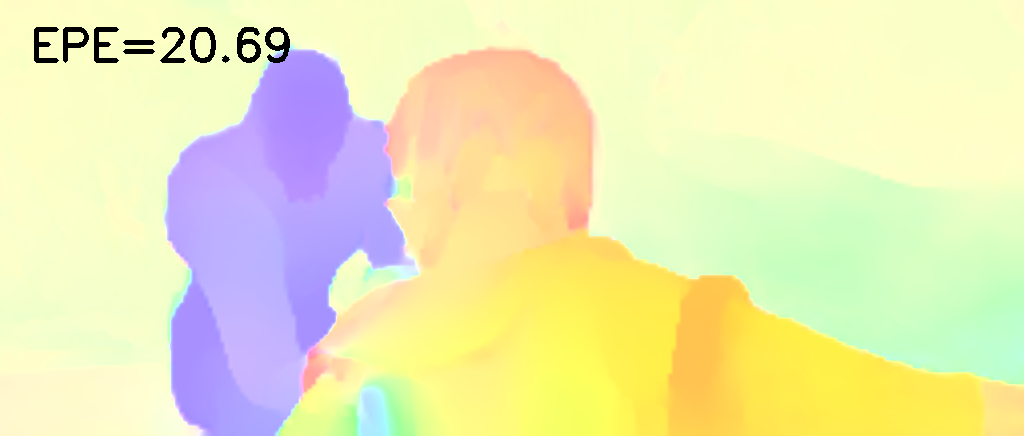}
			&
			\includegraphics[width=0.195\linewidth]{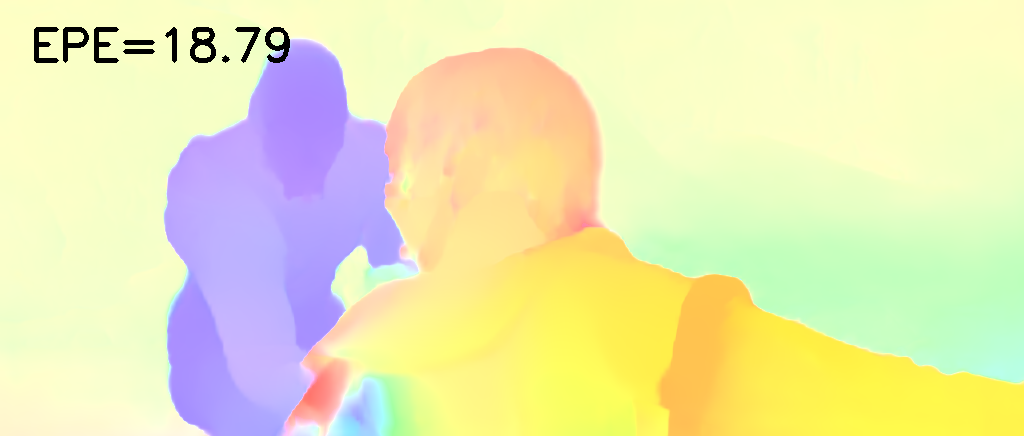}
			\vspace{-0.7mm}
			\\
			\includegraphics[width=0.195\linewidth]{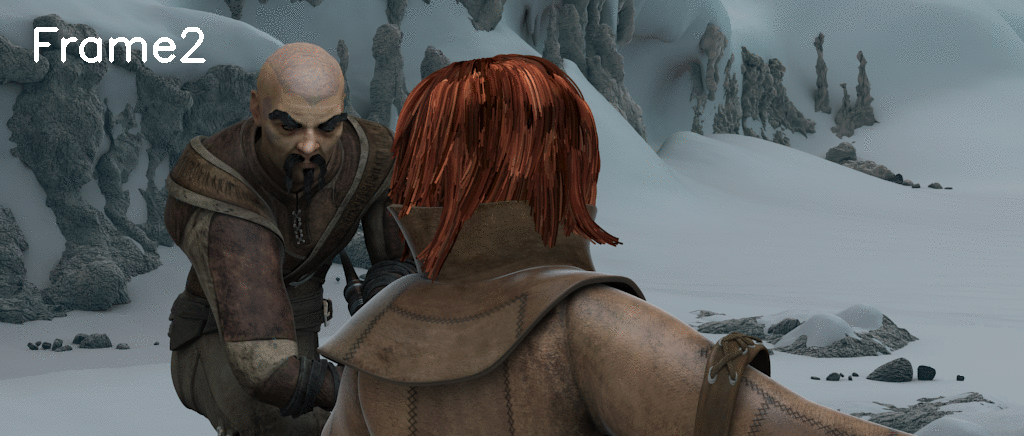}
			&
			\includegraphics[width=0.195\linewidth]{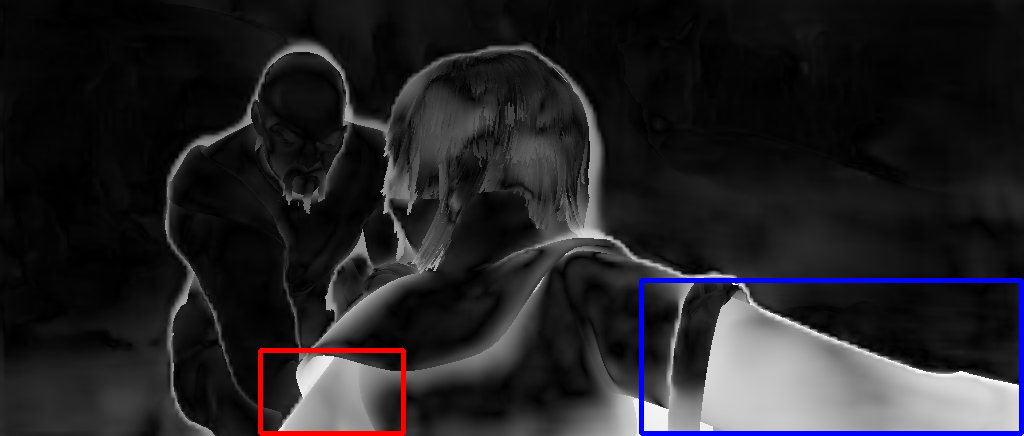}
			&
			\includegraphics[width=0.195\linewidth]{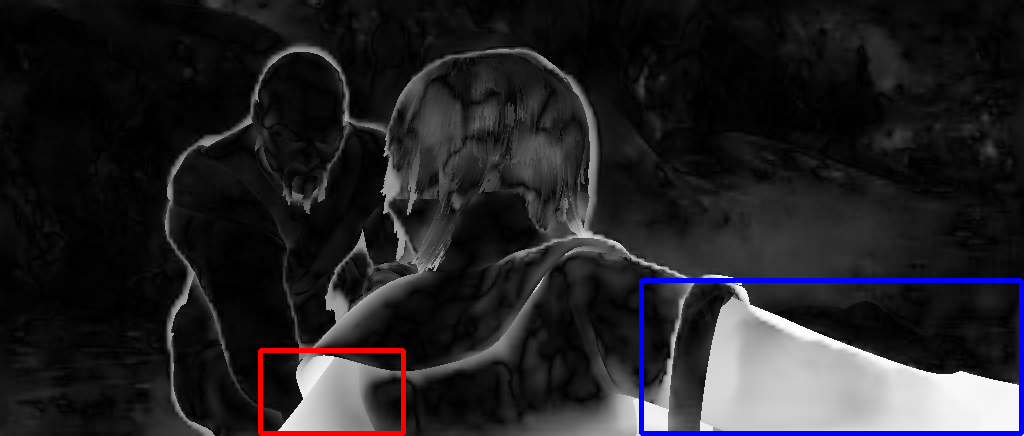}
			&
			\includegraphics[width=0.195\linewidth]{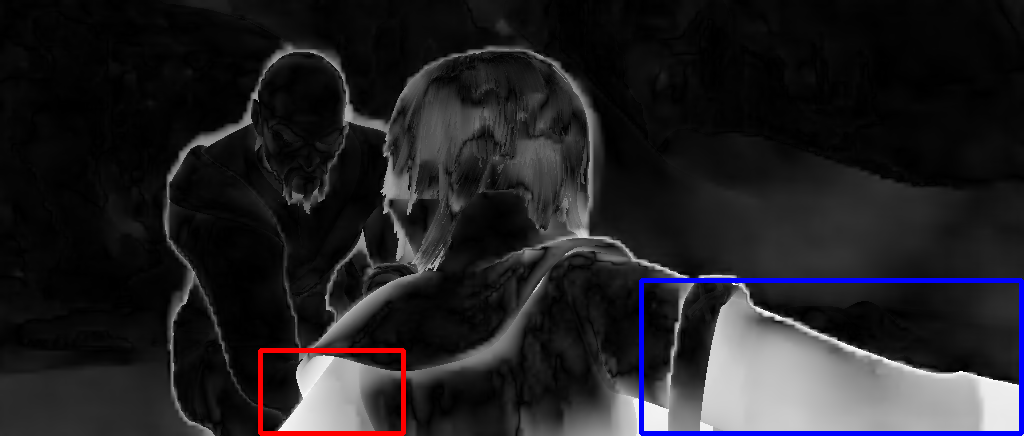}
			&
			\includegraphics[width=0.195\linewidth]{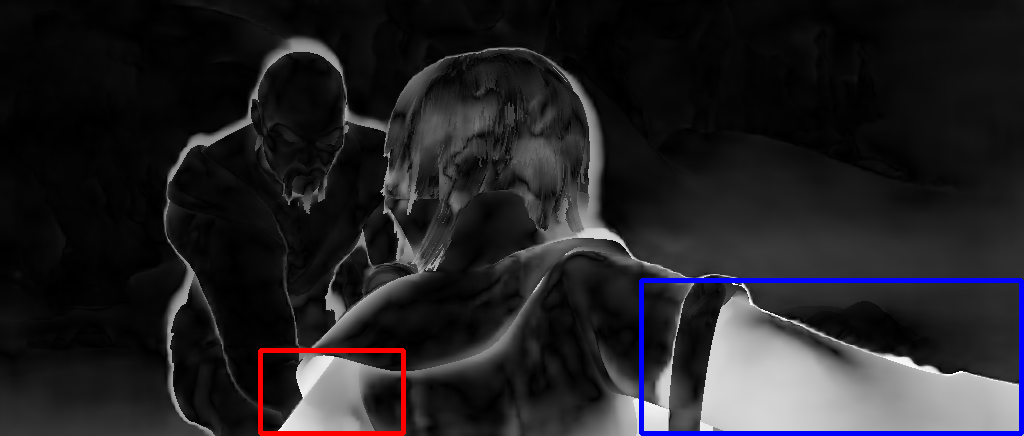}
			\vspace{-0.3mm}
			\\
			\includegraphics[width=0.195\linewidth]{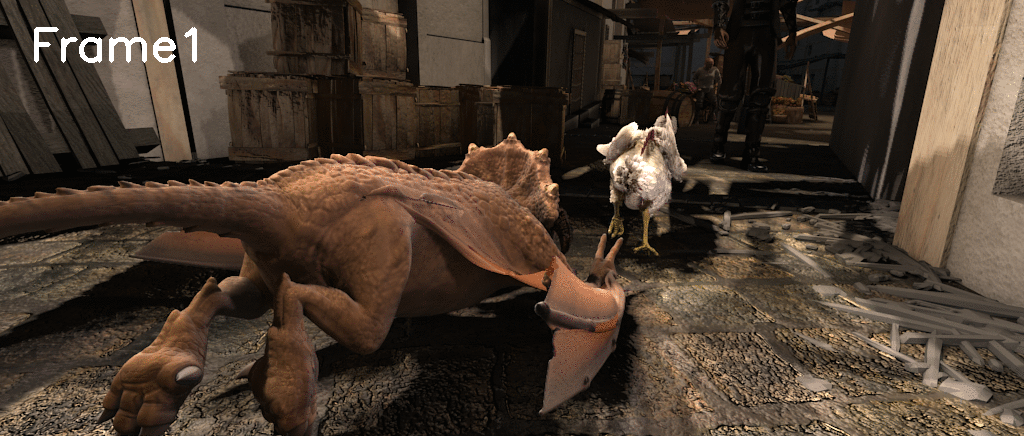}
			&
			\includegraphics[width=0.195\linewidth]{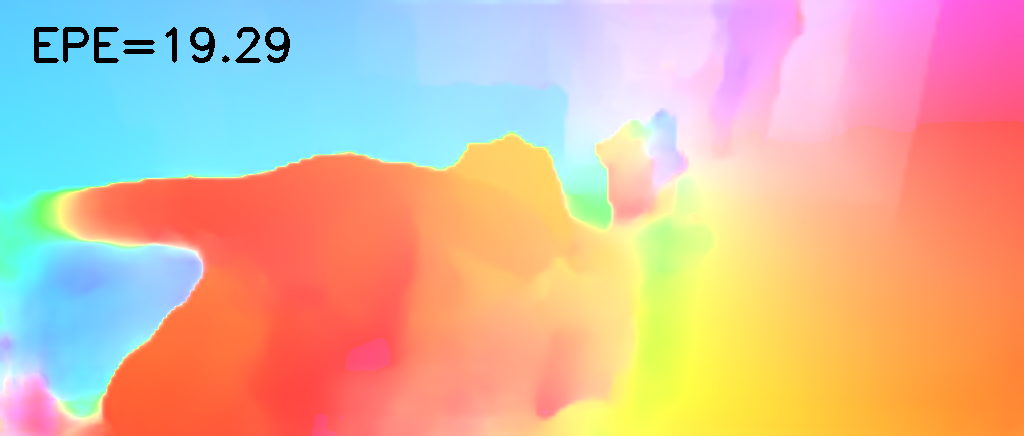}
			&
			\includegraphics[width=0.195\linewidth]{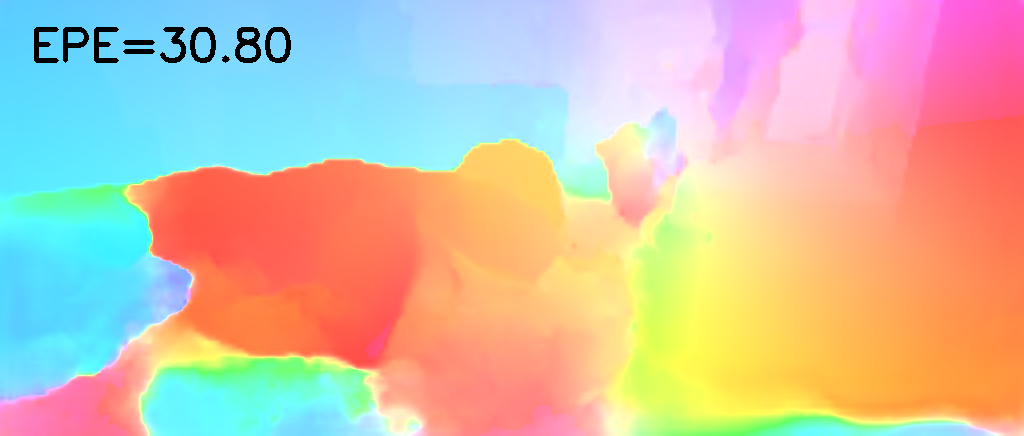}
			&
			\includegraphics[width=0.195\linewidth]{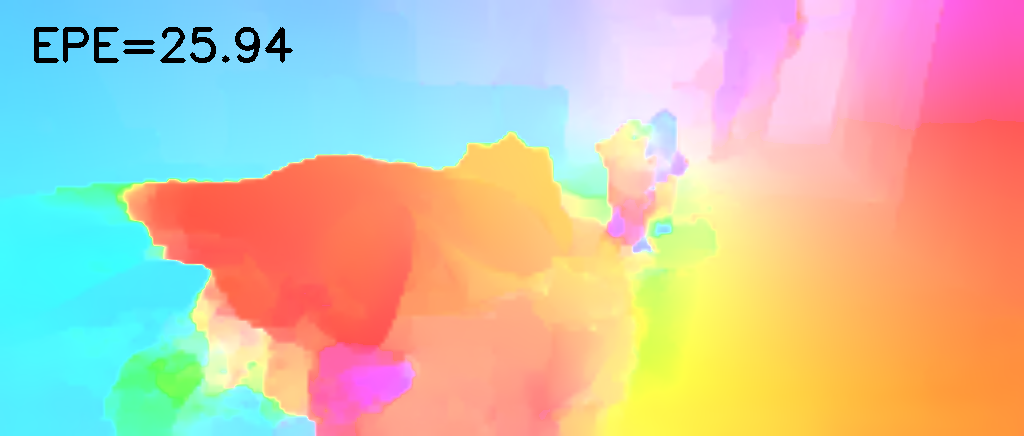}
			&
			\includegraphics[width=0.195\linewidth]{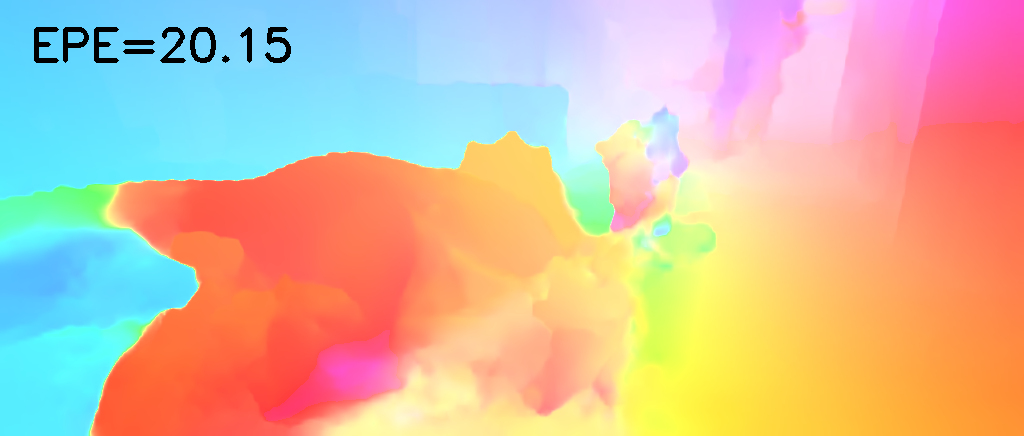}
			\vspace{-0.7mm}
			\\
			\includegraphics[width=0.195\linewidth]{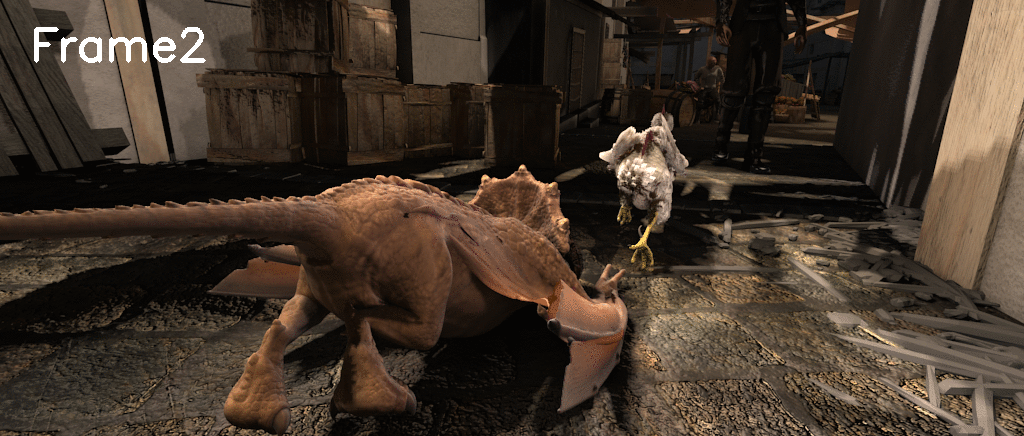}
			&
			\includegraphics[width=0.195\linewidth]{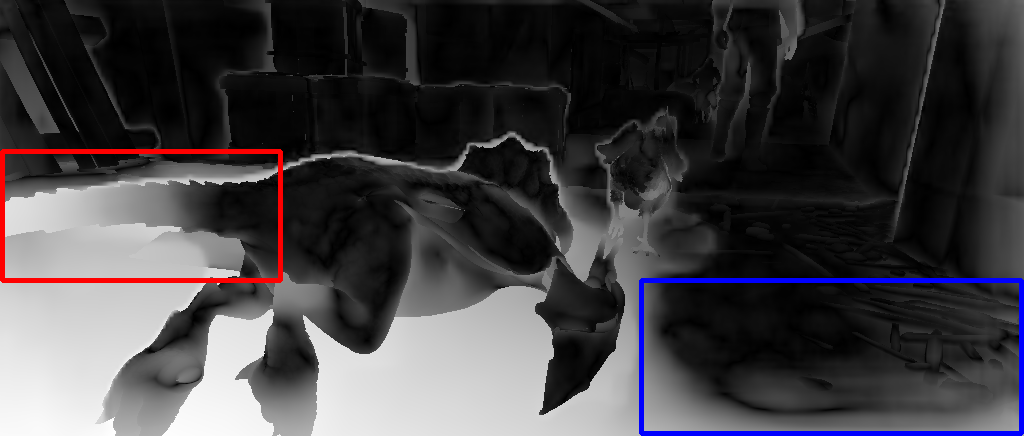}
			&
			\includegraphics[width=0.195\linewidth]{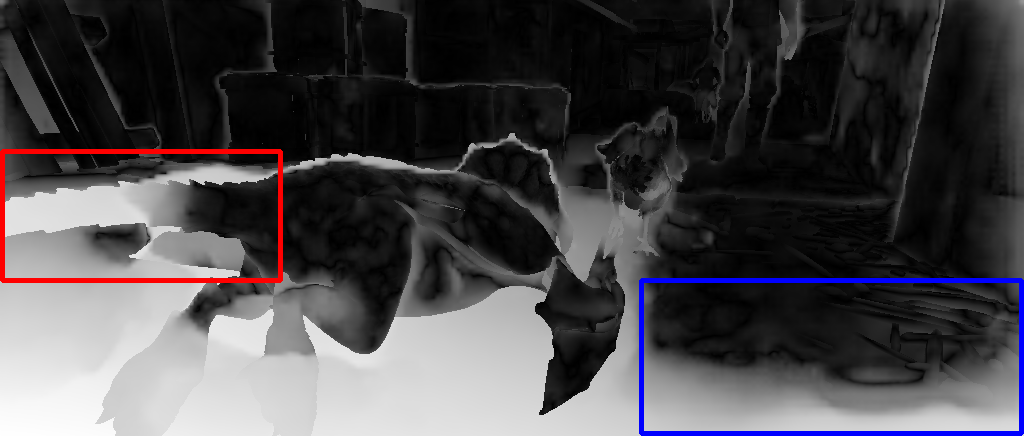}
			&
			\includegraphics[width=0.195\linewidth]{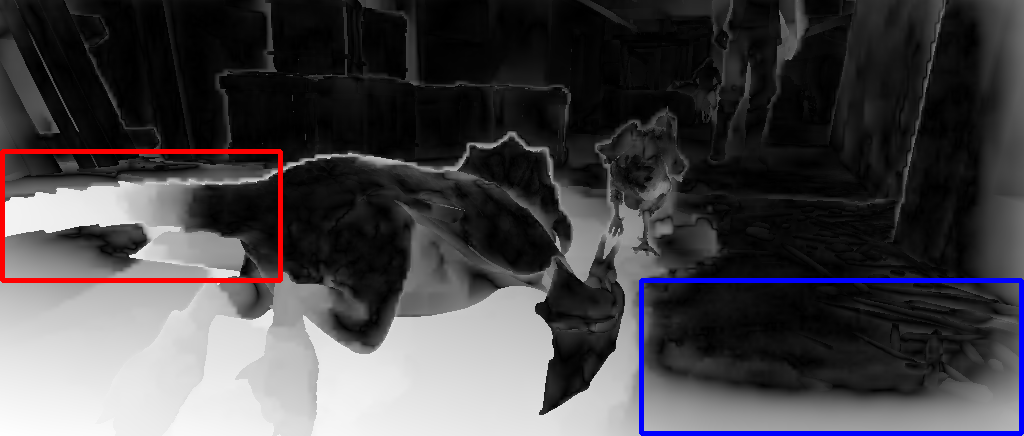}
			&
			\includegraphics[width=0.195\linewidth]{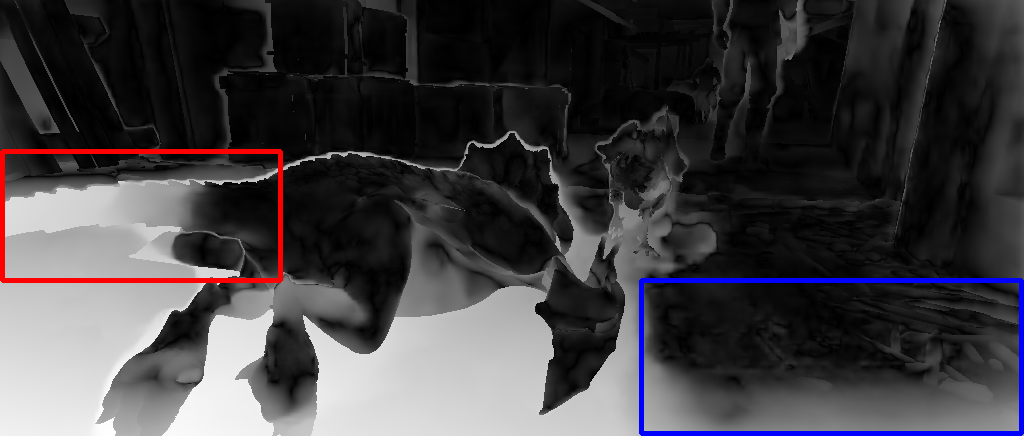}
			\vspace{-0.1mm}
			\\
			Input Frames & MDFlow & SimFlow~\cite{10.1007/978-3-030-58586-0_11} & UFlow~\cite{10.1007/978-3-030-58536-5_33} & UPFlow~\cite{Luo_2021_CVPR}
	\end{tabular}}
	\caption{\textbf{Qualitative results of state-of-the-art pyramid-based unsupervised optical flow methods on MPI Sintel test datasets.} Colored flow fields and flow error maps are interlaced. Fine details and less artifacts can be observed in the flow fields of proposed MDFlow. Zoom in for best view.}
	\label{fig:4}
\end{figure*}

\section{Experiments} \label{Experiments}
In this section, we first describe implementation details and training strategy of proposed MDFlow algorithm. Then, we quantitively and qualitatively compare our results with existing state-of-the-art approaches on standard benchmark datasets, including Flying Chairs~\cite{7410673}, MPI Sintel~\cite{Butler:ECCV:2012} and KITTI 2015~\cite{Menze2015CVPR}. Further, ablation study on proposed reliable matching selection mechanism and superior student architecture in forward distillation, as well as multi-target regularized learning in backward distillation are carried out. Finally, we show superior generalization ability of our method.

\begin{table}[t]
	\centering
	\renewcommand{\arraystretch}{1.1}
	\tabcolsep=2.4mm
	\caption{\textbf{Comparison of different optical flow architectures.} Running time and Computation complexity are measured on Sintel resolution images with one NVIDIA GTX 1080 Ti GPU.}
	\begin{tabular}{l|ccc}
		\toprule
		& Parameters (M) &  Runtime (ms) &  FLOPs (G) \\
		\midrule
		RAFT~\cite{teed2020raft} & 5.26 & 125 & 362.5 \\
		PWC-Net~\cite{8579029} & 8.75 & 34 & 90.8 \\
		FastFlowNet~\cite{Kong_2021_ICRA} & \textbf{1.37} & \textbf{11} & \textbf{12.2} \\
		\bottomrule
	\end{tabular}
	\label{tab:2}
\end{table}

\subsection{Implementation Detail} \label{Implementation Detail}
We employ the efficient PWC-Net~\cite{8579029} as teacher model, and the powerful RAFT~\cite{teed2020raft} as student model, whose model size, running time and computation complexity are compared in Table~\ref{tab:2}. It can be seen that even RAFT contains a little less parameters than PWC-Net, however, its running time and computation complexity is 4$\times$ larger. All our experiments are implemented in PyTorch and conducted on 2 NVIDIA Tesla V100 GPUs. We adopt the Adam~\cite{kingma2014method} and AdamW~\cite{loshchilov2018decoupled} optimizers to train the PWC-Net and RAFT respectively. Batch size is set to 4 for all experiments. Following previous work~\cite{Liu:2019:DDFlow,10.1007/978-3-030-58586-0_11,10.1007/978-3-030-58536-5_33}, we first pre-train MDFlow on Flying Chairs, and then fine-tune above teacher and student networks on each target dataset. As for Sintel, we use the combination of Clean and Final training parts, and adopt the fine-tuned teacher model for online evaluation. In regard to KITTI, we train on the KITTI 2015 multi-view extension training set, while removing frame numbers 9-12 in each sequence. Results of the teacher on KITTI 2015 test set are uploaded to KITTI website for online comparison. It takes overall about 4 days to train the three-stage MDFlow first on Chairs and then on Sintel or KITTI datasets, which also relys on the model complexity, IO speed and code parallelism degree.

\begin{table*}[t]
	\caption{\textbf{Quantitative results on Flying Chairs, MPI Sintel and KITTI 2015 datasets.} The metric EPE is the average endpoint error and Fl-all is the percentage of erroneous pixels over all pixels. Results in parentheses indicates it is evaluated using data it is trained on. The unavailable results are marked as `-'. `C', `T', `S' and `K' stand for Flying Chairs, FlyingThings3D, MPI Sintel and KITTI 2015 datasets respectively. Superscript `raw' means the raw part of corresponding datasets. $\rm K^{vo}$ means KITTI Visual Odometry dataset. (Stereo) denotes stereo data is used during training. For each item in supervised setting (top part), the best result is in \textbf{bold}. For each item in unsupervised setting (bottom part), the best result is \textcolor{red}{\textbf{boldfaced}}, and the second best is \textcolor{blue}{\underline{underlined}}.}
	\label{tab:3}
	\centering
	\renewcommand{\arraystretch}{1.1}
	\tabcolsep=4.4mm
	\begin{tabular}{llcccccccc}
		\toprule
		\multirow{2}{*}{Data} & \multirow{2}{*}{Method} & \multicolumn{1}{c}{C-test} & \multicolumn{2}{c}{S-train (EPE)} & \multicolumn{2}{c}{S-test (EPE)} & \multicolumn{1}{c}{K-15-train} & \multicolumn{1}{c}{K-15-test}\\
		\cmidrule(lr){3-3}
		\cmidrule(lr){4-5}
		\cmidrule(lr){6-7}
		\cmidrule(lr){8-8}
		\cmidrule(lr){9-9}
		& & EPE & Clean & Final & Clean & Final & EPE & Fl-all \\
		\midrule
		$\rm C+T$ & FlowNet2~\cite{8099662} & - & 2.02 & 3.14 & \textbf{3.96} & \textbf{6.02} & 10.06 & - \\
		$\rm S/K$ & FlowNet2-ft~\cite{8099662} & - & (1.45) & (2.01) & 4.16 & 5.74 & (2.3) & 10.41\% \\
		$\rm C+T$ & FastFlowNet~\cite{Kong_2021_ICRA} & - & 2.89 & 4.14 & - & - & 12.24 & - \\
		$\rm S/K$ & FastFlowNet-ft~\cite{Kong_2021_ICRA} & - & (2.08) & (2.71) & 4.89 & 6.08 & (2.13) & 11.22\% \\
		$\rm C+T$ & PWC-Net~\cite{8579029} & \textbf{2.30} & 2.55 & 3.93 & - & - & 10.35 & - \\ 
		$\rm S/K$ & PWC-Net-ft~\cite{8579029} & - & (1.70) & (2.21) & 3.86 & 5.13 & (2.16) & 9.60\% \\
		$\rm C+T$ & RAFT~\cite{teed2020raft} & - & \textbf{1.43} & \textbf{2.71} & - & - & \textbf{5.04} & - \\
		$\rm S/K$ & RAFT-ft~\cite{teed2020raft} & - & (\textbf{0.77}) & (\textbf{1.20}) & \textbf{2.08} & \textbf{3.41} & (\textbf{1.5}) & \textbf{5.27\%} \\
		\midrule
		$\rm SY/K^{raw}$ & UnFlow-CSS~\cite{Meister:2018:UUL} & - & - & 7.91 & - & 10.22 & 8.10 & - \\
		$\rm C+S/K$ & OccAwareFlow~\cite{8578611} & 3.30 & (4.03) & (5.95) & 7.95 & 9.15 & 8.88 & 31.20\% \\
		$\rm R+S/K$ & MFOccFlow~\cite{Janai_2018_ECCV} & - & (3.89) & (5.52) & 7.23 & 8.81 & 6.59 & 22.94\% \\
		$\rm C+S/K$ & EPIFlow~\cite{8953885} & - & (3.54) & (4.99) & 7.00 & 8.51 & 5.56 & 16.95\% \\
		$\rm C+S/K$ & DDFlow~\cite{Liu:2019:DDFlow} & 2.97 & (2.92) & (3.98) & 6.18 & 7.40 & 5.72 & 14.29\% \\
		$\rm S^{raw}/K$ & SelFlow~\cite{Liu:2019:SelFlow} & - & (2.88) & (3.87) & 6.56 & 6.57 & 4.84 & 14.19\% \\
		$\rm C+S/K$ & STFlow~\cite{9201360} & \textcolor{blue}{\underline{2.53}} & (2.91) & (3.59) & 6.12 & 6.63 & 3.56 & 13.83\% \\
		$\rm K^{raw} (Stereo)$ & UnOS~\cite{Wang_2019_CVPR} & - & - & - & - & - & 5.58 & 18.00\% \\
		$\rm K (Stereo)$ & Flow2Stereo~\cite{Flow2Stereo} & - & - & - & - & - & 3.54 & 11.10\% \\
		$\rm S^{raw}/K^{raw}$ & ARFlow~\cite{Liu_2020_CVPR} & - & 2.79 & 3.73 & 4.78 & 5.89 & 2.85 & 11.80\% \\
		$\rm C+S/K$ & SimFlow~\cite{10.1007/978-3-030-58586-0_11} & 2.69 & (2.86) & (3.57) & 5.92 & 6.92 & 5.19 & 13.38\% \\
		$\rm C+S/K$ & UFlow~\cite{10.1007/978-3-030-58536-5_33} & 2.55 & (2.50) & (3.39) & 5.21 & 6.50 & 2.71 & 11.13\% \\
		$\rm S+S^{raw}/K^{raw}$ & DistillFlow~\cite{9444870} & - & (2.61) & (3.70) & 4.23 & 5.81 & 2.93 & 10.54\% \\
		$\rm C+S/K$ & OIFlow~\cite{9477059} & \textcolor{blue}{\underline{2.53}} & (2.44) & (3.35) & 4.26 & 5.71 & 2.57 & 9.81\% \\
		$\rm S/K^{raw}$ & ASFlow~\cite{9625946} & - & (2.40) & \textcolor{blue}{\underline{(2.89)}} & 4.56 & 5.86 & 2.47 & 9.67\% \\
		$\rm S^{raw}/K^{raw}$ & CoT-AMFlow~\cite{Wang_CoRL_2020} & - & - & - & \textcolor{red}{\textbf{3.96}} & \textcolor{red}{\textbf{5.14}} & - & 10.34\% \\
		$\rm K^{vo}+K (Stereo)$ & FLC~\cite{Chi_2021_CVPR} & - & - & - & - & - & \textcolor{red}{\textbf{2.35}} & 9.70\% \\
		$\rm S/K^{raw}$ & UPFlow~\cite{Luo_2021_CVPR} & - & \textcolor{blue}{\underline{(2.33)}} & \textcolor{red}{\textbf{(2.67)}} & 4.68 & \textcolor{blue}{\underline{5.32}} & \textcolor{blue}{\underline{2.45}} & \textcolor{blue}{\underline{9.38\%}} \\
		$\rm C+S/K$ & MDFlow-Fast (Ours) & 2.75 & (2.53) & (3.47) & 4.73 & 5.99 & 3.02 & 11.43\% \\
		$\rm C+S/K$ & MDFlow (Ours) & \textcolor{red}{\textbf{2.48}} & \textcolor{red}{\textbf{(2.17)}} & (3.14) & \textcolor{blue}{\underline{4.16}} & 5.46 & \textcolor{blue}{\underline{2.45}} & \textcolor{red}{\textbf{8.91\%}} \\
		\bottomrule
	\end{tabular}
\end{table*}

\subsection{Training Strategy} \label{Training Strategy}
On each dataset, we perform proposed three-stage training pipeline for mutual knowledge distillation, where the first and last stage both take 300k iterations to train the teacher, while the second stage takes 100k iterations to train the student. When training on Flying Chairs, learning rate is initially set to $1e-4$ and decays half at 100k, 150k, 200k and 250k iterations in both stage 1 and stage 3. As for stage 2, it is initially set to $4e-4$ and decays half at 20k, 40k, 60k and 80k iterations. When fine-tuning on Sintel and KITTI, we use the same training schedule as Chairs, however, learning rate is divided by 2 in all corresponding phases. Across stages, both the teacher and the student load latest updated checkpoint for initialization, and are random initialized from scratch before training on Chairs. The data augmentor $\mathcal{A}$ includes operations of random flipping, resizing, rotating, cropping and color jittering. According to hyperparameter search, weight coefficients of $\lambda_1, \lambda_2, \lambda_3, \lambda_4$ in Eq.~\ref{eq:4} and Eq.~\ref{eq:7} are respectively set to $1.0, 0.05, 0.1, 0.1$ for Chairs and Sintel, while being $5.0, 0.05, 0.02, 0.1$ for KITTI. Note that the smoothness coefficient $\lambda_1$ in KITTI is 5 $\times$ of that in Chairs and Sintel in the first stage, while the photometric coefficient $\lambda_3$ in KITTI is only $1/5$ of that in Chairs and Sintel in the third stage. It is because that KITTI uses the second order smoothness, a larger coefficient can reach a reasonable regularization strength in the first stage. And in the third stage, the non-occluded region in KITTI benefits more from the distillation term rather than the noisy reconstruction term.

\subsection{Comparison to State-of-the-art Methods}
\subsubsection{Quantitative Results}
As shown in Table~\ref{tab:3}, we quantitatively compare MDFlow with existing supervised and unsupervised methods on three leading optical flow benchmarks with standard metrics, where our approach outperforms most pyramid-based unsupervised methods on almost all datasets. The current state-of-the-art SMURF~\cite{Stone_2021_CVPR} uses the unfair recurrent model during evaluation, which is not listed in Table~\ref{tab:3} because of large computation and time delay. Moreover, proposed MDFlow framework is general and does not depend on particular flow architecture and unsupervised initialization. Due to the diverse training configuration used in previous work, we list their training data in the first column for reference.

On Flying Chairs, MDFlow reduces previous best EPE from 2.53 to 2.48 slightly, possibly because of the relative simple motion in this domain. As for more challenging Sintel and KITTI 2015 datasets, our method achieves obvious improvement over others. On Sintel training, we obtains the best EPE of 2.17 on the Clean pass, and only behaves a little worse than UPFlow~\cite{Luo_2021_CVPR} and ASFlow~\cite{9625946} on the Final pass. For the online test benchmarks, our method achieves EPE of 4.16 and 5.46 on Clean and Final respectively, even outperforming DistillFlow~\cite{9444870} (4.23 and 5.81), which is trained on Sintel raw dataset, containing the unfair test image sequences. The method CoT-AMFlow~\cite{Wang_CoRL_2020} behaves best on both Sintel Clean and Final test sets. We attribute the reason to be that it both employs a more complex adaptive modulation flow network and uses the unfair Sintel raw movie sequences.

\begin{figure*}[t]
	\footnotesize
	\centering
	\resizebox{1.0\textwidth}{!}{
		\begin{tabular}{@{}c @{\hskip 0.01in} c @{\hskip 0.01in} c @{\hskip 0.01in} c @{\hskip 0.01in} c@{}}
			\includegraphics[width=0.195\linewidth]{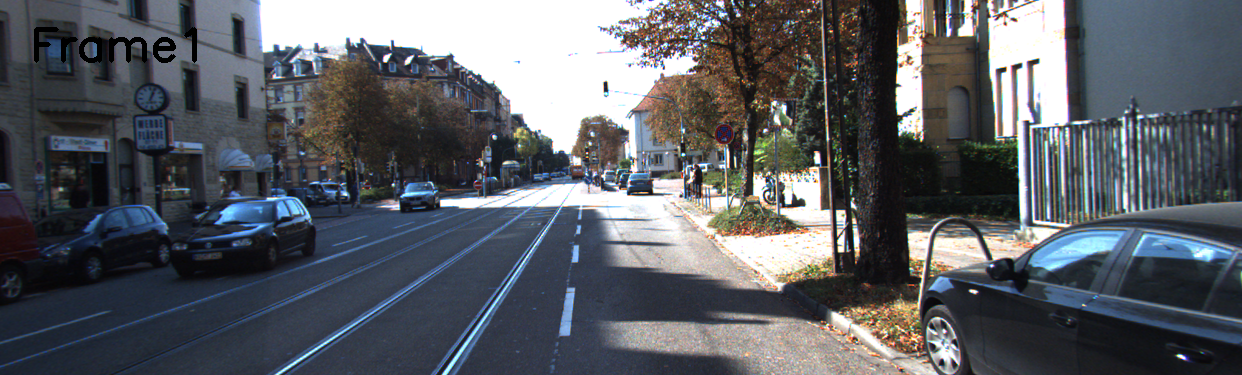}
			&
			\includegraphics[width=0.195\linewidth]{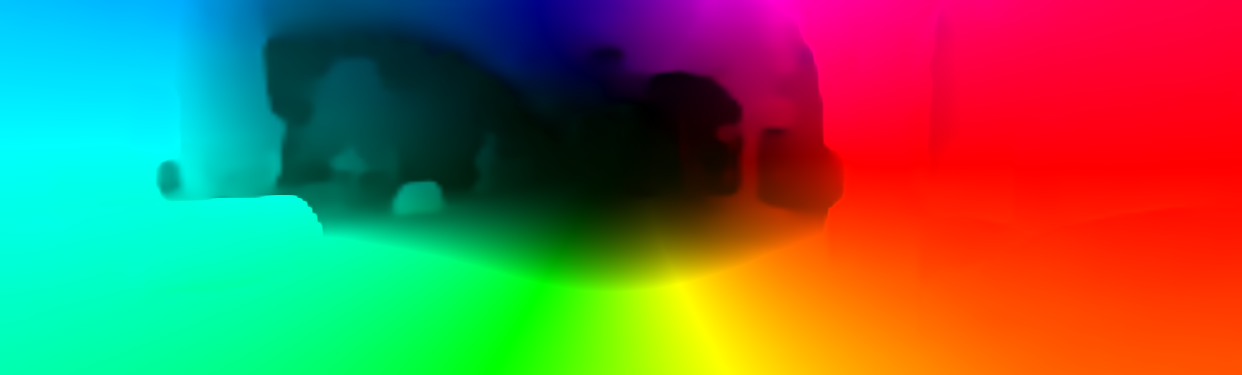}
			&
			\includegraphics[width=0.195\linewidth]{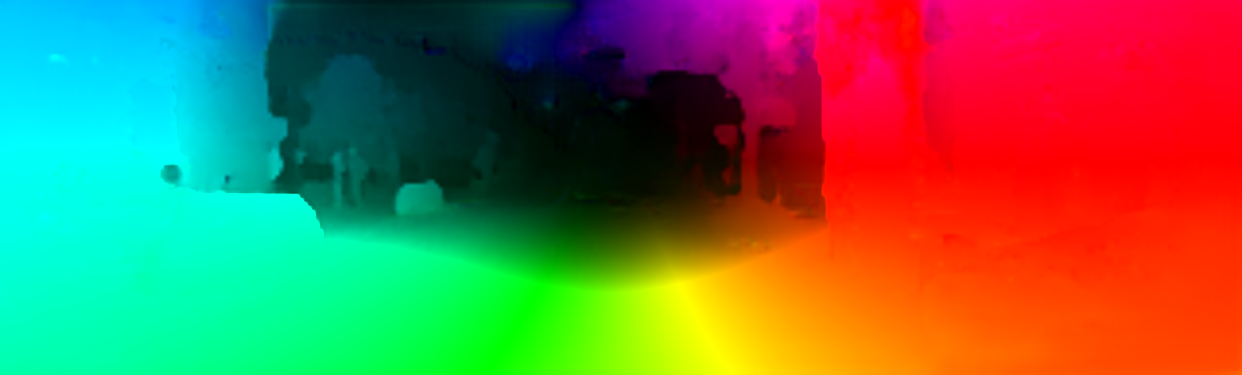}
			&
			\includegraphics[width=0.195\linewidth]{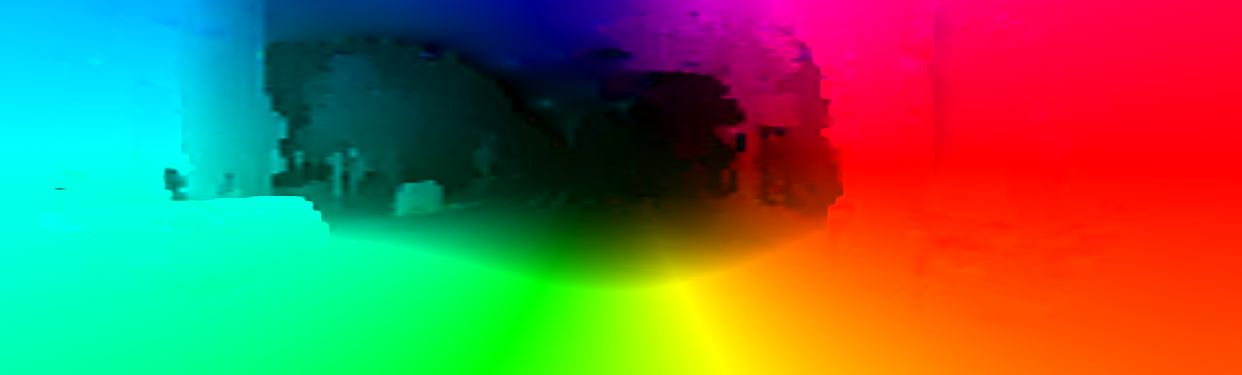}
			&
			\includegraphics[width=0.195\linewidth]{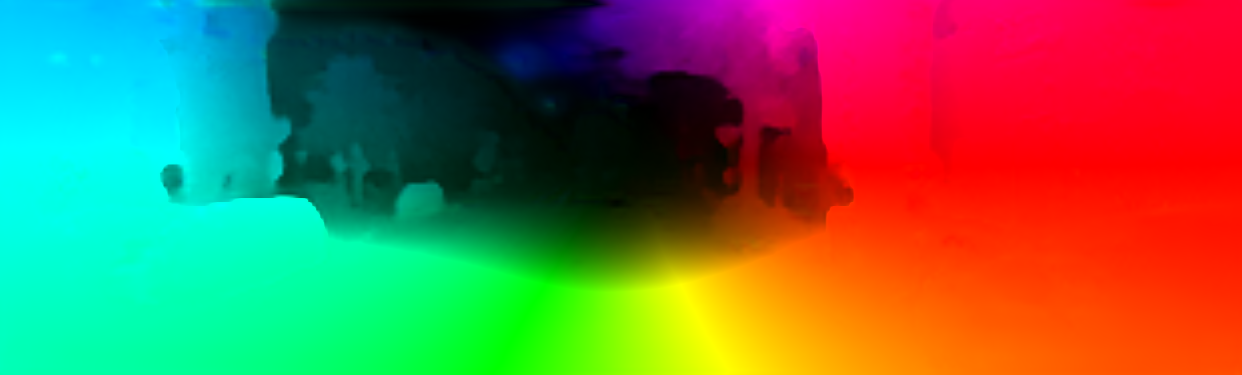}
			\vspace{-0.7mm}
			\\
			\includegraphics[width=0.195\linewidth]{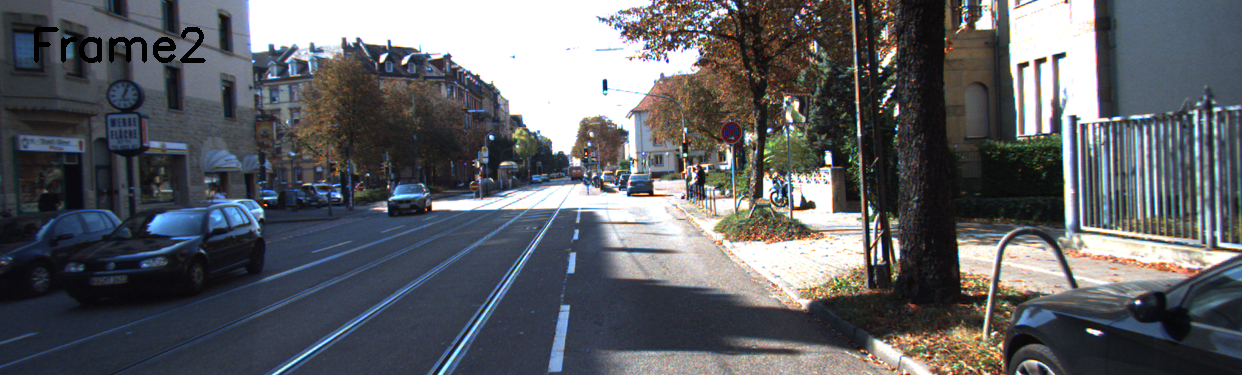}
			&
			\includegraphics[width=0.195\linewidth]{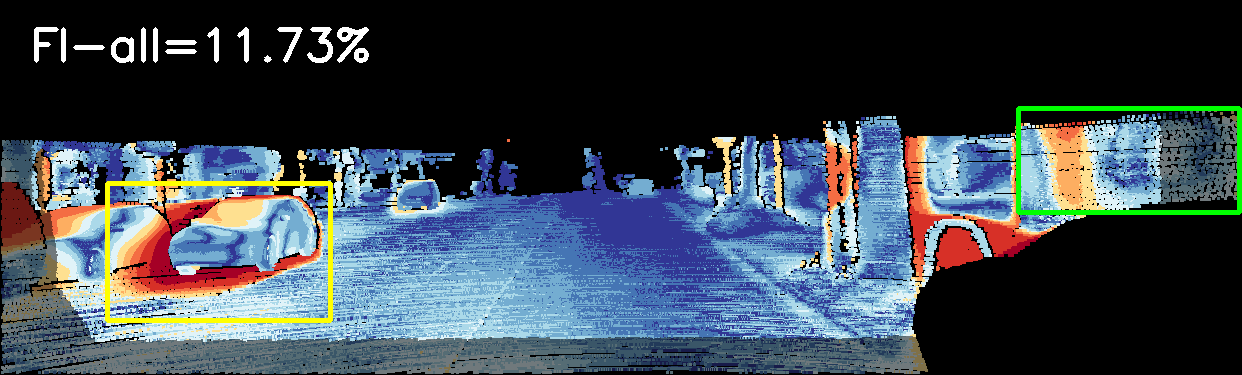}
			&
			\includegraphics[width=0.195\linewidth]{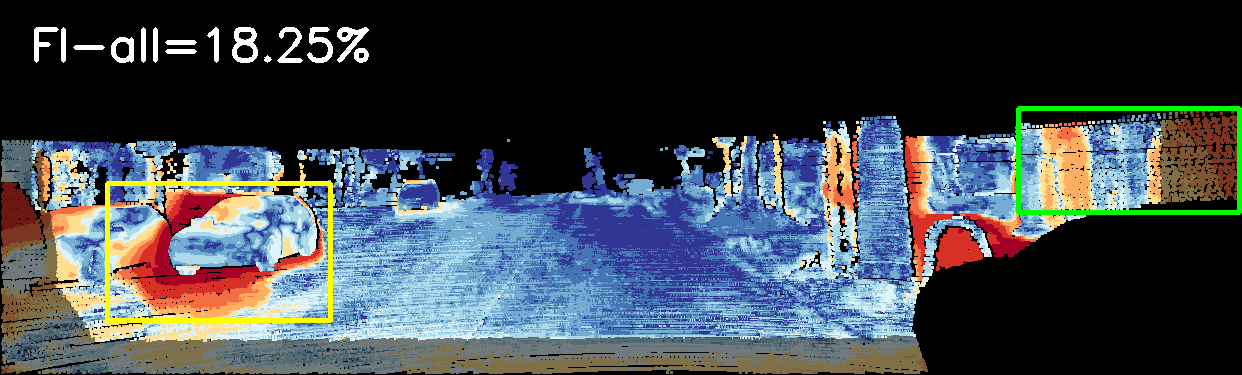}
			&
			\includegraphics[width=0.195\linewidth]{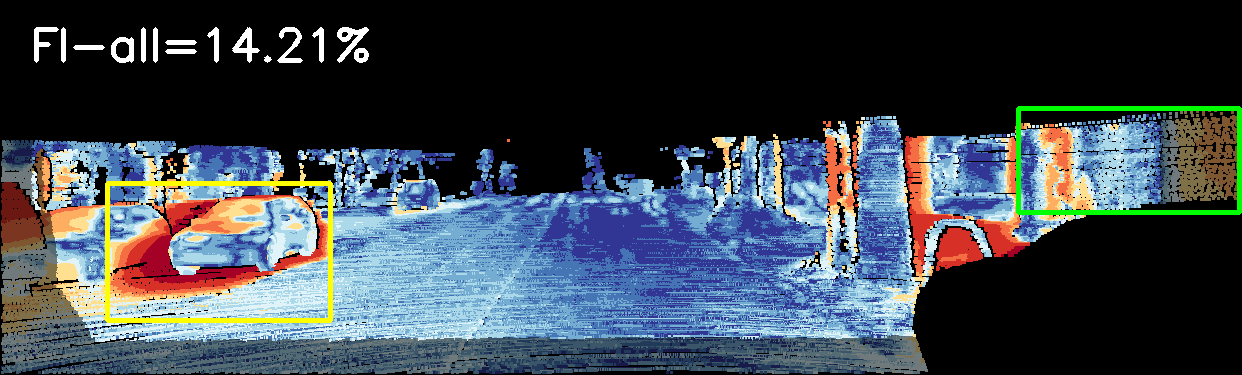}
			&
			\includegraphics[width=0.195\linewidth]{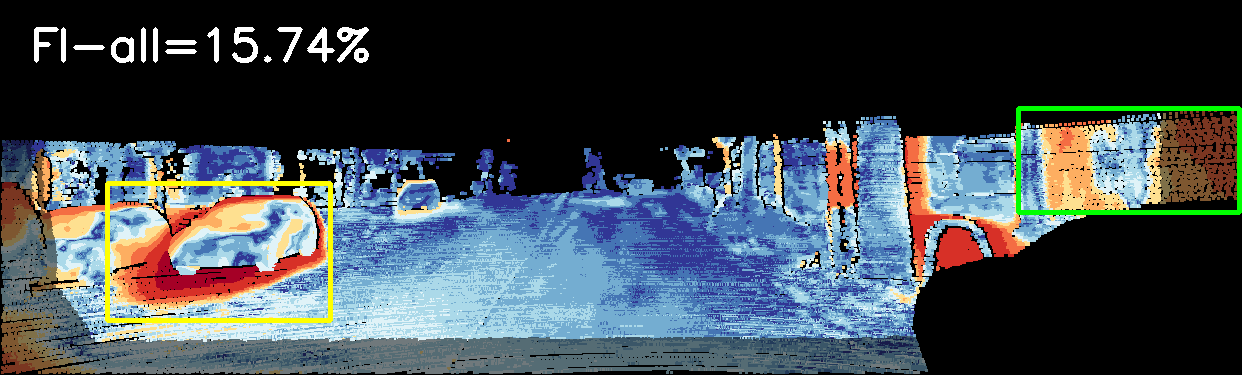}
			\vspace{-0.3mm}
			\\
			\includegraphics[width=0.195\linewidth]{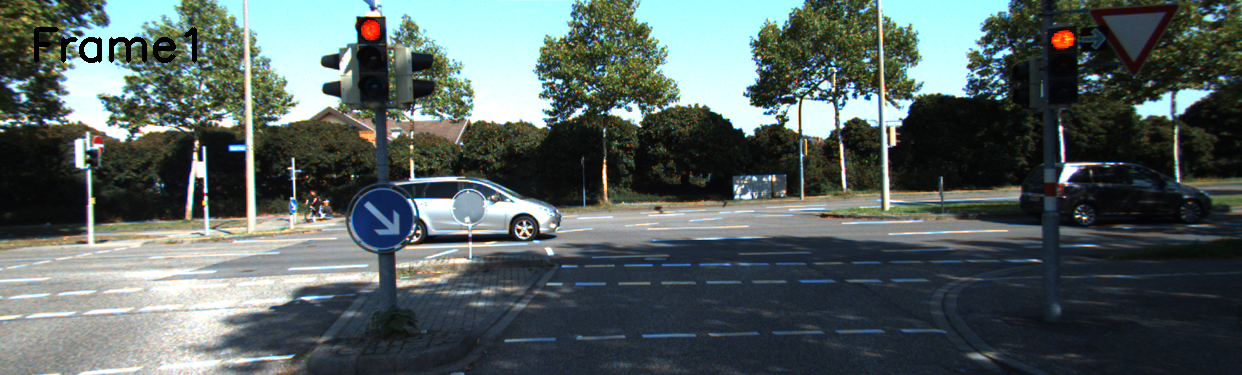}
			&
			\includegraphics[width=0.195\linewidth]{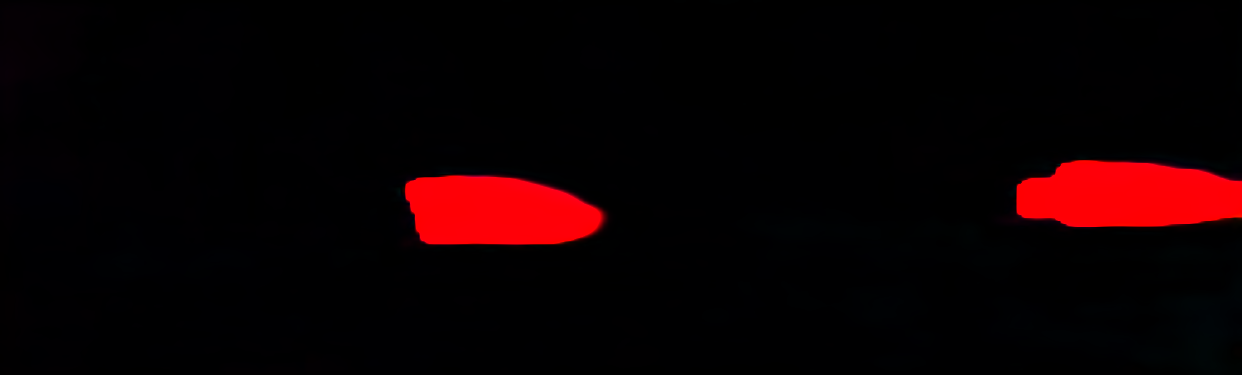}
			&
			\includegraphics[width=0.195\linewidth]{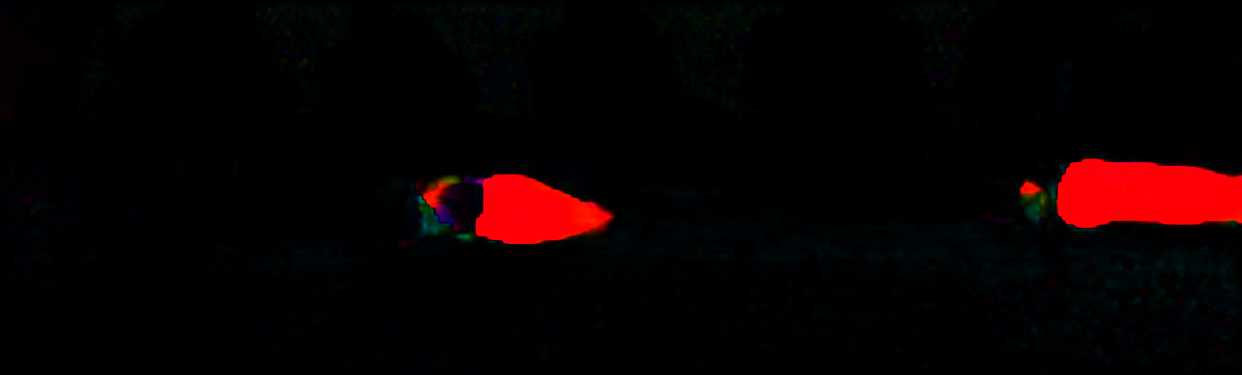}
			&
			\includegraphics[width=0.195\linewidth]{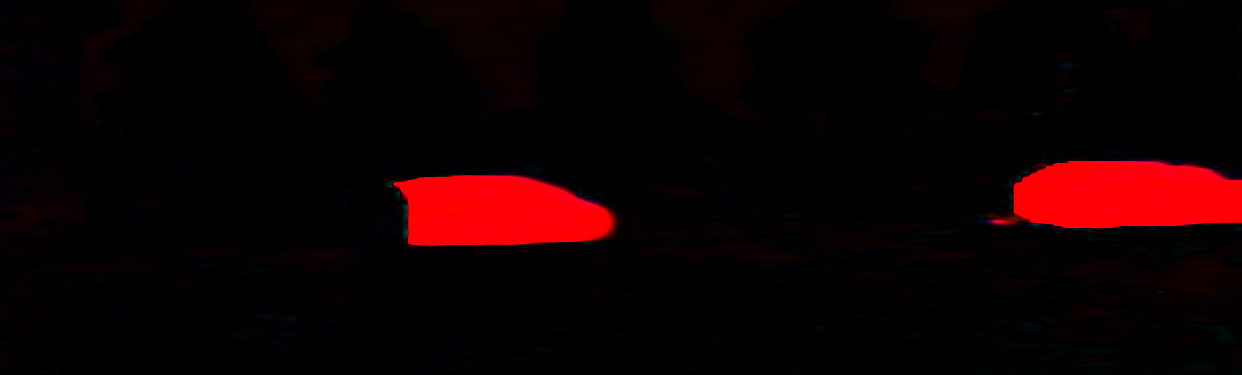}
			&
			\includegraphics[width=0.195\linewidth]{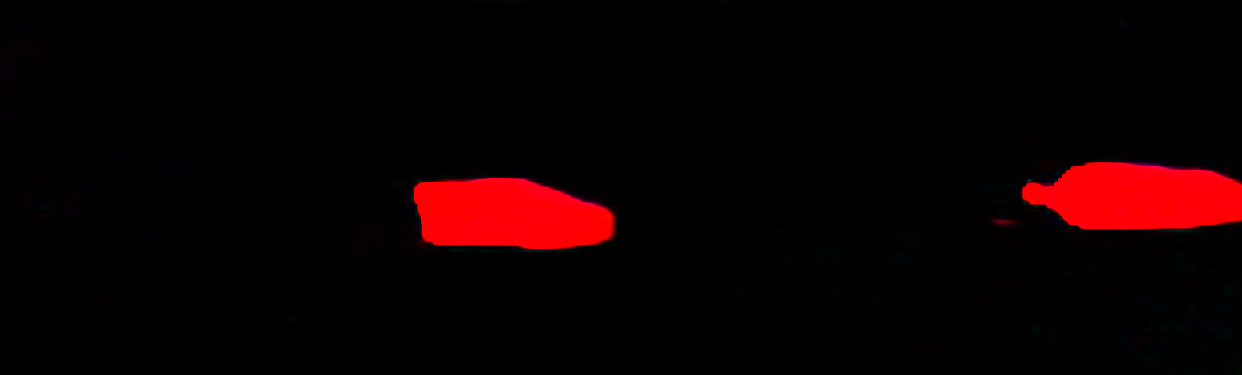}
			\vspace{-0.7mm}
			\\
			\includegraphics[width=0.195\linewidth]{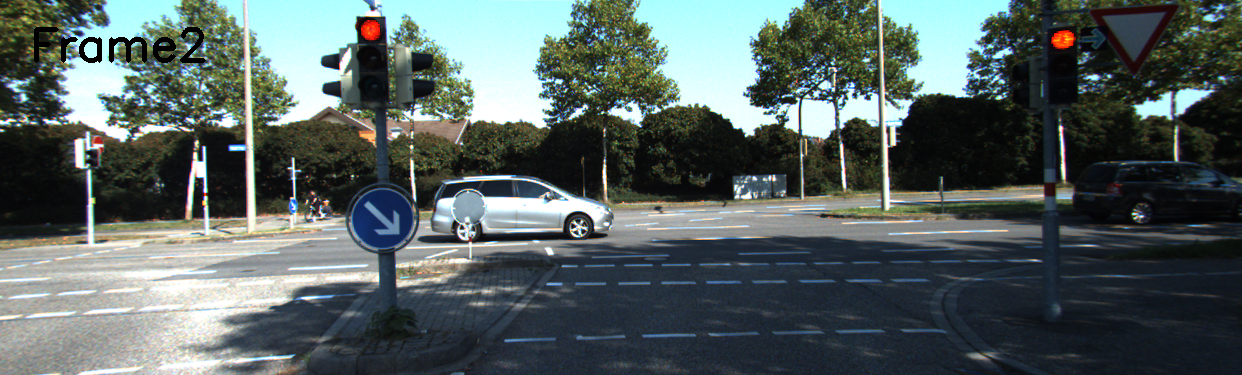}
			&
			\includegraphics[width=0.195\linewidth]{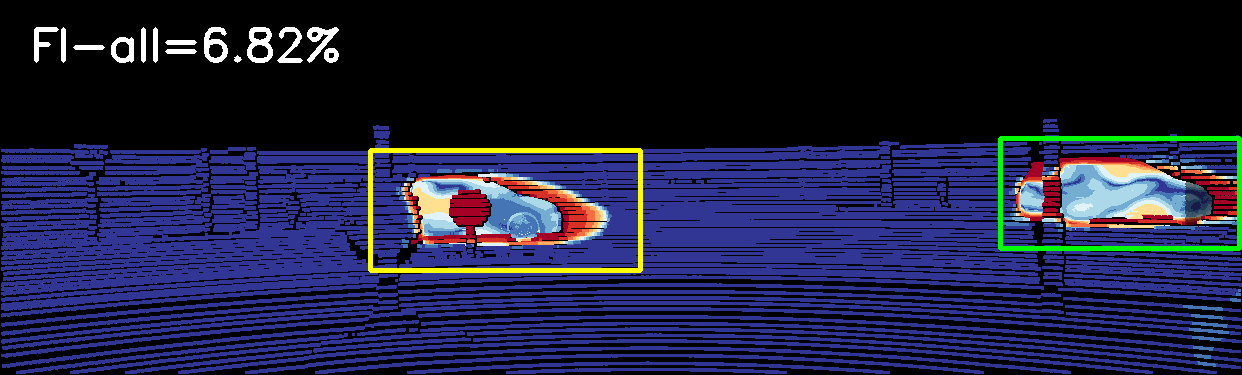}
			&
			\includegraphics[width=0.195\linewidth]{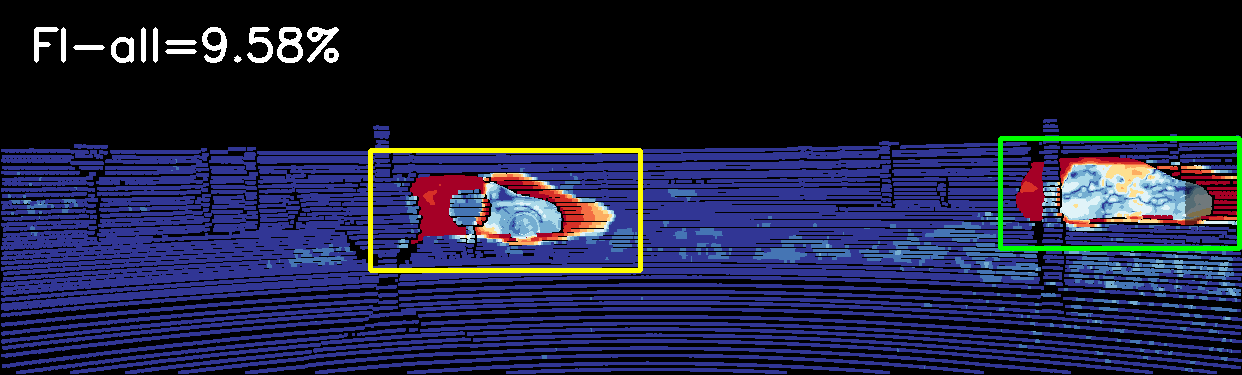}
			&
			\includegraphics[width=0.195\linewidth]{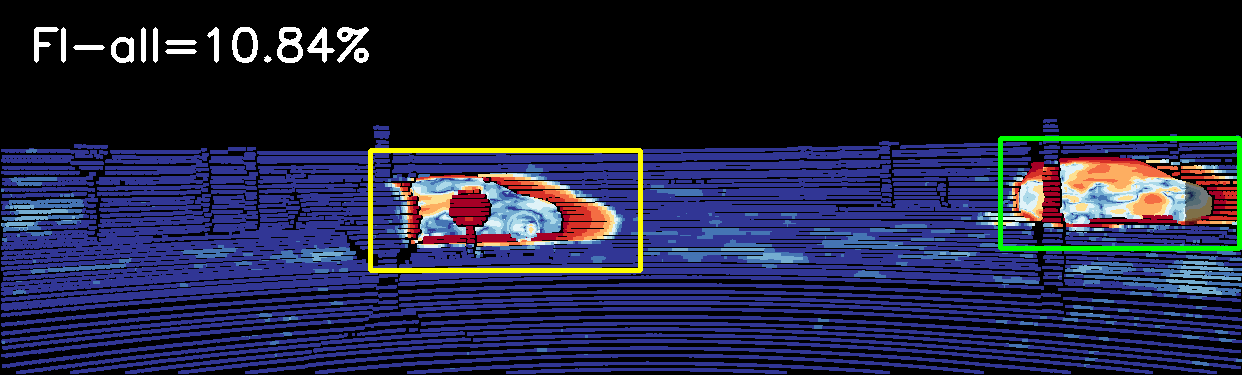}
			&
			\includegraphics[width=0.195\linewidth]{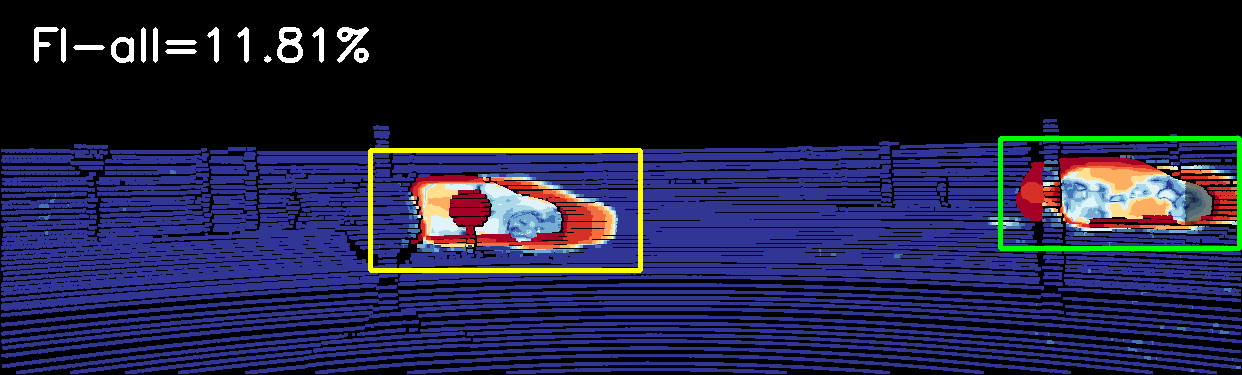}
			\vspace{-0.3mm}
			\\
			\includegraphics[width=0.195\linewidth]{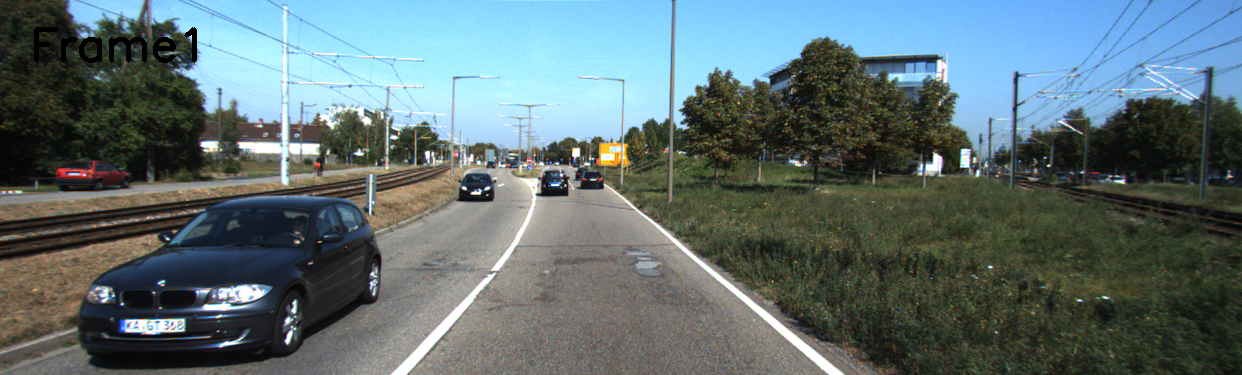}
			&
			\includegraphics[width=0.195\linewidth]{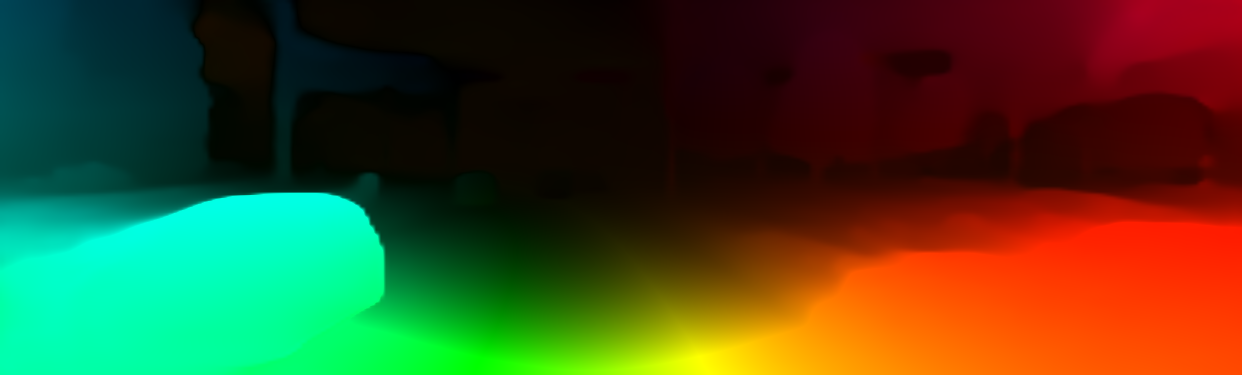}
			&
			\includegraphics[width=0.195\linewidth]{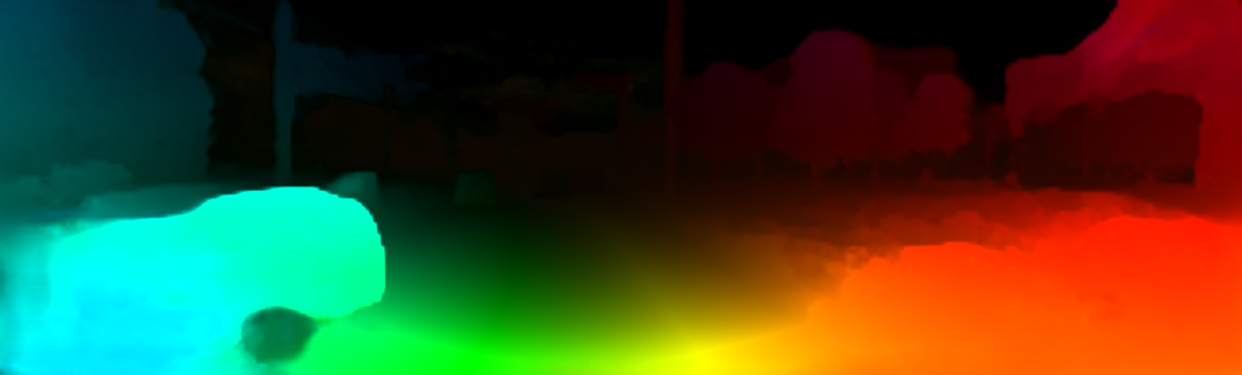}
			&
			\includegraphics[width=0.195\linewidth]{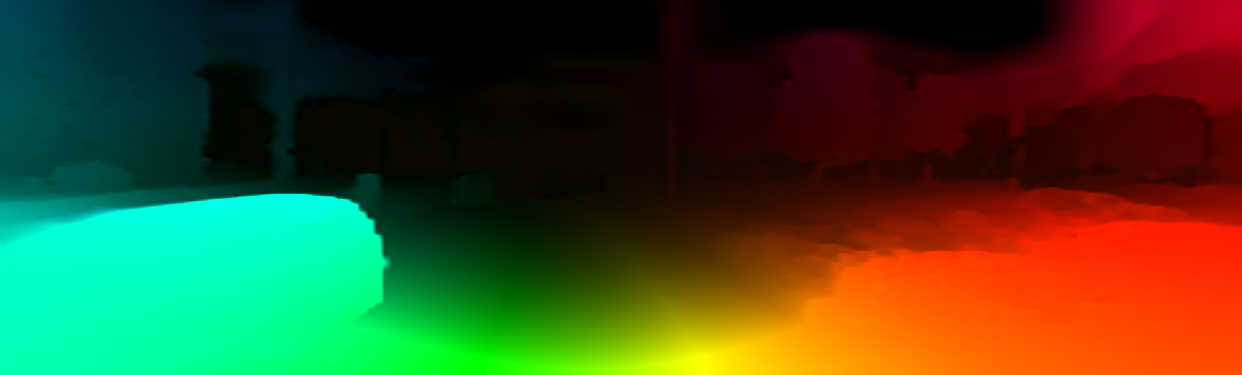}
			&
			\includegraphics[width=0.195\linewidth]{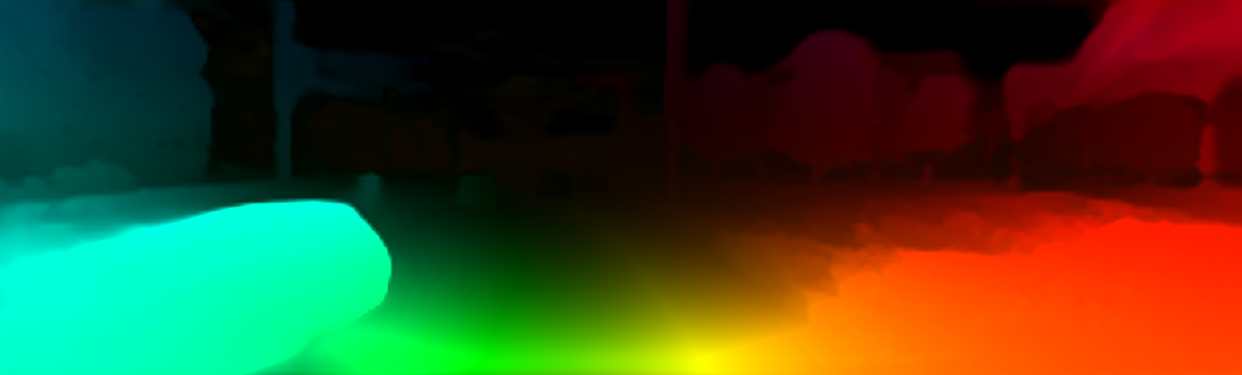}
			\vspace{-0.7mm}
			\\
			\includegraphics[width=0.195\linewidth]{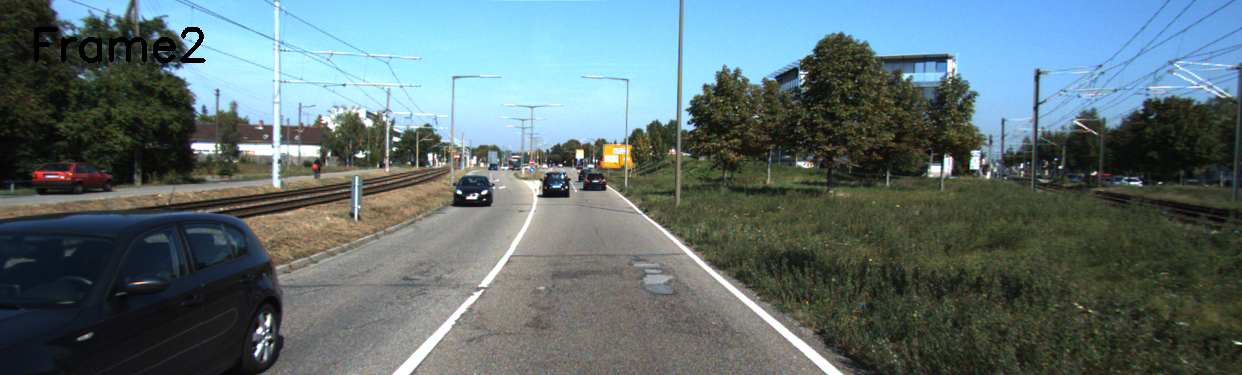}
			&
			\includegraphics[width=0.195\linewidth]{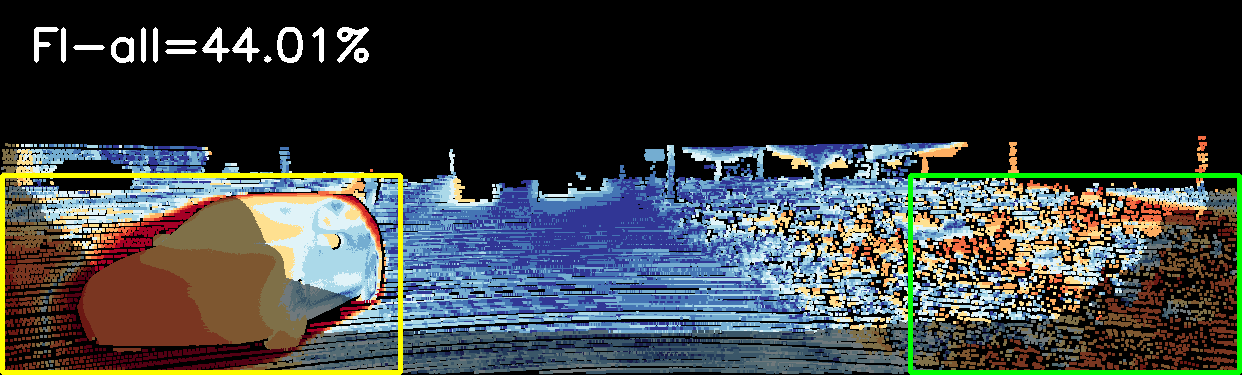}
			&
			\includegraphics[width=0.195\linewidth]{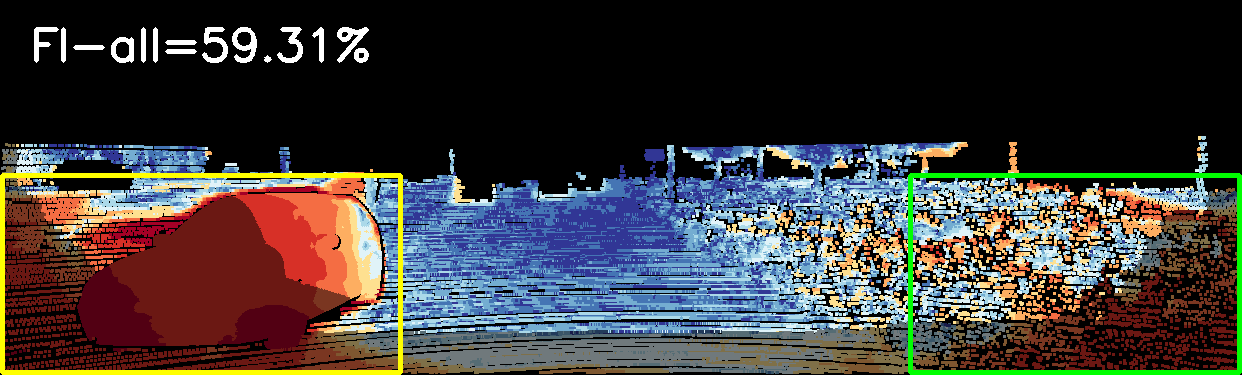}
			&
			\includegraphics[width=0.195\linewidth]{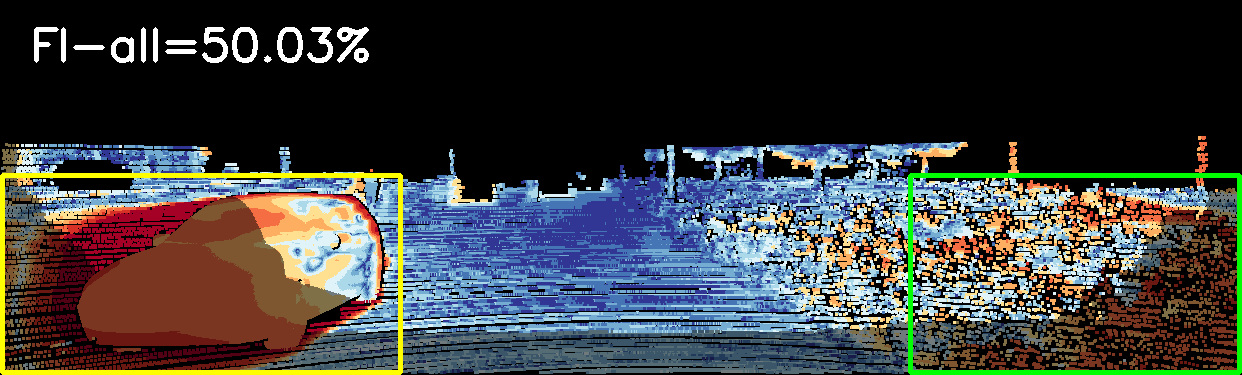}
			&
			\includegraphics[width=0.195\linewidth]{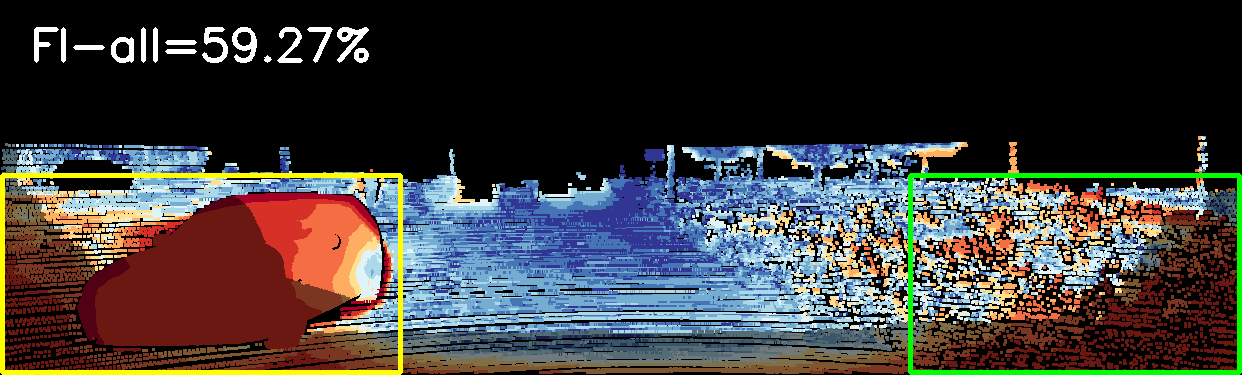}
			\vspace{-0.1mm}
			\\
			Input Frames & MDFlow & SimFlow~\cite{10.1007/978-3-030-58586-0_11} & UFlow~\cite{10.1007/978-3-030-58536-5_33} & SelFlow~\cite{Liu:2019:SelFlow}
	\end{tabular}}
	\caption{\textbf{Qualitative results of state-of-the-art pyramid-based unsupervised optical flow methods on KITTI 2015 test datasets.} Colored flow fields and flow error maps are interlaced. Fine details and less artifacts can be observed in the flow fields of proposed MDFlow. Zoom in for best view.}
	\label{fig:5}
\end{figure*}

In regard to KITTI 2015, we achieve the same second best EPE of 2.45 with UPFlow~\cite{Luo_2021_CVPR} on the training set, but without additional parameters and inference delay, which is 9.2\% improvement over comparable UFlow~\cite{10.1007/978-3-030-58536-5_33} (2.71). We further upload flow predictions of the efficient teacher on test images to the KITTI benchmark website and achieve 8.91\% Fl-all error rate, outperforming the state-of-the-art pyramid-based unsupervised approach UPFlow~\cite{Luo_2021_CVPR}, \textit{i.e.}, 8.91\% \textit{vs} 9.38\%. It is worth noting that proposed MDFlow also behaves better than the state-of-the-art binocular SLAM method FLC~\cite{Chi_2021_CVPR}, that shows the room for improvement in this research area.

To facilitate a large number of downstream video tasks, especially for embedded real-time applications, we replace the orignal PWC-Net~\cite{8579029} with a more efficient FastFlowNet~\cite{Kong_2021_ICRA} as the teacher model, and repeat the same experiments as before. Model size, running time and computation complexity of FastFlowNet are compared in Table~\ref{tab:2}. As shown in Table~\ref{tab:3}, MDFlow-Fast can approach the state-of-the-art unsupervised flow accuracy while only containing 1.37 M parameters and consuming much less computation and inference time, demonstrating the efficient flow architecture of FastFlowNet and the efficient learning framework of MDFlow.

\subsubsection{Qualitative Results}
Some qualitative results on MPI Sintel and KITTI 2015 test datasets are depicted in Figure~\ref{fig:4} and Figure~\ref{fig:5} for visual comparison respectively. It can be seen that flow fields of proposed MDFlow can keep better motion scene structure layout than other methods, especially on regions with large occlusion and complex texture.

\subsection{Ablation Study}
To verify the effectiveness of proposed contributions, we conduct thorough ablation experiments of reliable mutual knowledge distillation algorithm considering all training stages and different training settings, and our findings are reported in Table~\ref{tab:4}. All ablation experiments are trained on Sintel training or KITTI 2015 multi-view extension datasets with corresponding loss functions in their training stage. We first report results of unsupervised initialization (stage 1) for teacher model on Sintel training and KITTI 2015 training sets using Eq.~\ref{eq:4} as the baseline, which is denoted as E1 in Table~\ref{tab:4}.

\begin{table*}[t]
	\caption{\textbf{Ablation study on MPI Sintel and KITTI 2015 datasets.} $\rightarrow$ and $\leftarrow$ represent forward and backward distillation processes, and no arrow means using a single model without distillation. `Init' stands for checkpoints that corresponding models are initialized with. `C' means checkpoints pre-trained on Flying Chairs dataset. $RR$ stands for removal rate in Figure~\ref{fig:3}. $\mathcal{L}_{sup}, \mathcal{L}_{ph}, \mathcal{L}_{sm}$ refer to different loss functions in Eq. 7. Results are tested on the model which is optimized during this ablation experiment.}
	\label{tab:4}
	\centering
	\renewcommand{\arraystretch}{1.1}
	\tabcolsep=5.0mm
	\begin{tabular}{cllllcccccccc}
		\toprule
		\multirow{2}{*}{Stage} & \multirow{2}{*}{ID} & \multirow{2}{*}{Mode} & \multirow{2}{*}{Init} & \multirow{2}{*}{Setting} & \multicolumn{2}{c}{S-train (EPE)} & \multicolumn{2}{c}{K-15-train} \\
		\cmidrule(lr){6-7}
		\cmidrule(lr){8-9}
		& & & & & Clean & Final & EPE & Fl-all \\
		\midrule
		\multirow{2}{*}{1} & E1 & PWC & C & 300k Iterations & (2.39) & (3.34) & 2.66 & 9.64\% \\
		& E2 & PWC & C & 700k Iterations & (2.38) & (3.32) & 2.64 & 9.66\% \\
		\midrule
		\multirow{6}{*}{2} & E3 & \multirow{4}{*}{PWC$\rightarrow$PWC} & \multirow{4}{*}{E1$\rightarrow$C} & $RR=0\%$ & (2.43) & (3.33) & 2.64 & 8.86\% \\
		& E4 & & & $RR=10\%$ & (2.32) & (3.28) & 2.54 & 8.50\% \\
		& E5 & & & $RR=20\%$ & (2.36) & (3.31) & 2.60 & 8.48\% \\
		& E6 & & & $RR=30\%$ & (2.37) & (3.35) & 2.67 & 8.60\% \\
		& E7 & PWC & E1 & $RR=10\%$ & (2.33) & (3.31) & 3.00 & 9.89\% \\
		& E8 & PWC$\rightarrow$RAFT & E1$\rightarrow$C & $RR=10\%$ & (\textbf{2.16}) & (\textbf{3.15}) & \textbf{2.31} & \textbf{8.31\%} \\
		\hline
		\multirow{3}{*}{3}  & E9 & PWC$\leftarrow$PWC & E1$\leftarrow$E4 & $\mathcal{L}_{sup}$ & (2.31) & (3.25) & 2.53 & 8.36\% \\
		& E10 & PWC$\leftarrow$RAFT & E1$\leftarrow$E8 & $\mathcal{L}_{sup}$ & (2.27) & (3.22) & 2.48 & 8.22\% \\
		& E11 & PWC$\leftarrow$RAFT & E1$\leftarrow$E8 & $\mathcal{L}_{sup}+\mathcal{L}_{ph}+\mathcal{L}_{sm}$ & (\textbf{2.17}) & (\textbf{3.14}) & \textbf{2.45} & \textbf{8.10\%} \\
		\midrule
		4 & E12 & PWC$\rightarrow$RAFT & E11$\rightarrow$E8 & $RR=10\%$ & (2.08) & (3.10) & 2.20 & 7.85\% \\
		\midrule
		5 & E13 & PWC$\leftarrow$RAFT & E11$\leftarrow$E12 & $\mathcal{L}_{sup}+\mathcal{L}_{ph}+\mathcal{L}_{sm}$ & (2.16) & (3.14) & 2.36 & 8.04\% \\
		\bottomrule
	\end{tabular}
\end{table*}

\subsubsection{Reliable Matching Selection}
As shown in the middle part of Table~\ref{tab:4}, we conduct an ablation to explore the effectiveness of proposed confidence matching selection mechanism. According to Figure~\ref{fig:3}, removal rate is set to four typical values as $0\%, 10\%, 20\%$ and $30\%$, denoted by E3, E4, E5 and E6 in Table~\ref{tab:4}, where $0\%$ means that original non-occluded map is adopted as valid mask during forward knowledge distillation. In this group, both the teacher and the student models are instantiated as PWC-Net. Compared with the baseline E1, forward distillation with $RR=0\%$ has almost no improvement on the student, due to the mismatched regions in pseudo labels. When we remove $10\%$ prediction of the teacher according to proposed confidence ranking mechanism in Eq.~\ref{eq:5}, the student yields the lowest endpoint error on both Sintel and KITTI 2015 training sets. As more uncertain labels being removed, the EPE gets larger, while the Fl-all metric on KITTI 2015 behaves the best with $RR=20\%$. Therefore, we use $RR=10\%$ in MDFlow, that is equivalent to adopting $\tau$ of 3.22 and 3.59 on Sintel and KITTI respectively in Eq.~\ref{eq:5} according to Table~\ref{tab:1}. Speak of here, one may ask whether the distillation procedure is necessary, since we can directly apply proposed confidence mask to $\mathcal{L}_{sup}(\mathcal{T} | \mathcal{T}, \mathcal{A})$ when optimizing Eq.~\ref{eq:4}. To explore this, we further carry out an experiment named `PWC' in stage 2 marked as E7 in Table~\ref{tab:4}. As can be seen, it can approach the result of distillation counterpart on Sintel, while the results behave much worse on KITTI, whose reason may be the instability when combining the sparse valid mask with unsupervised loss in challenging scenes.

\subsubsection{Stronger Student Model}
Thanks to the decouple nature of our mutual distillation framework, we can employ any optical flow architecture as student without worrying about introducing additional cost in real deployment. Inspired by recent success on supervised optical flow task, for the first time, we try to explore whether a stronger student model can achieve better results when training on pseudo labels with the same amount of noise. As expected, when we employ advanced RAFT~\cite{teed2020raft} as the student, endpoint error is reduced by $6.9\%$, $4.0\%$ and $9.1\%$ on Sintel Clean, Final and KITTI 2015 respectively, that is shown in E8 of stage 2. Significantly improved results demonstrate the effectiveness of proposed reliable forward knowledge distillation process. Moreover, E9 is the corresponding backward distillation experiment to E4, which jointly constitute a complete mutual distillation procedure but do not employ a stronger student. Compare E9 and E10, we can conclude that a stronger student can improve the final performance of the teacher.

\subsubsection{Multi-Target Backward Distillation}
Superior accuracy in stage 2 is at the cost of inference delay, thus transferring the better student knowledge back to the efficient teacher model makes sense. One can directly regard student's prediction as pseudo labels, and train the teacher in a supervised manner, like what has been done in stage 2. However, as E10 shown in Table~\ref{tab:4}, performance of final teacher model declines back distinctly compared with E8, whose reason may be the limited learning capability of efficient network and relatively difficult task due to strong data augmentation. To deal with this problem, we formulate this step as a multi-target learning pipeline by introducing an unsupervised objective for regularization. As M11 listed in Table~\ref{tab:4}, proposed approach gets better results than pure supervision counterpart, and can well maintain or even surpass the powerful student model. Furthermore, we do sensitivity analysis of the photometric loss $\mathcal{L}_{ph}$ and the smoothness loss $\mathcal{L}_{sm}$ in backward distillation stage according to Eq.~\ref{eq:7}, where original $\lambda_3 \mathcal{L}_{ph}$ and $\lambda_4 \mathcal{L}_{sm}$ are multiplied by a scale factor of $\alpha$ and $\beta$ respectively. As shown in Table~\ref{tab:6}, $\alpha$ and $\beta$ are alternately set to 0.5 and 1.5 to perform a perturbation for weighting hyperparameters $\lambda_3$ and $\lambda_4$. It can be seen that a relatively large coefficient for $\mathcal{L}_{ph}$ is more benefit on Sintel, while a relatively large coefficient for $\mathcal{L}_{sm}$ is more helpful on KITTI. The reason may be Sintel dataset includes more non-rigid motion, while brightness noise is more obvious in real world KITTI dataset. In summary, in a reasonable range of values for $\lambda_3$ and $\lambda_4$, mutli-target backward distillation behaves better than the pure supervision counterpart, demonstrating the robustness of proposed knowledge distillation approach.

\begin{table}[t]
	\caption{\textbf{Sensitivity analysis of the photometric and the smoothness objective functions in the backward distillation stage.}}
	\label{tab:6}
	\centering
	\renewcommand{\arraystretch}{1.1}
	\tabcolsep=4.0mm
	\begin{tabular}{ccccc}
		\toprule
		$\mathcal{L}_{sup} + \alpha\lambda_3 \mathcal{L}_{ph}$ & \multicolumn{2}{c}{S-train (EPE)} & \multicolumn{2}{c}{K-15-train} \\
		\cmidrule(lr){2-3}
		\cmidrule(lr){4-5}
		$ + \beta \lambda_4 \mathcal{L}_{sm}$ & Clean & Final & EPE & Fl-all \\
		\midrule
		$\alpha=1.0, \, \beta=1.0$ & (\textbf{2.17}) & (\textbf{3.14}) & 2.45 & \textbf{8.10\%} \\
		$\alpha=0.5, \, \beta=1.0$ & (2.21) & (3.18) & 2.46 & 8.13\% \\
		$\alpha=1.5, \, \beta=1.0$ & (2.18) & (3.16) & 2.48 & 8.16\% \\
		$\alpha=1.0, \, \beta=0.5$ & (2.19) & (3.17) & 2.49 & 8.21\% \\
		$\alpha=1.0, \, \beta=1.5$ & (2.23) & (3.20) & \textbf{2.43} & 8.12\% \\
		\bottomrule
	\end{tabular}
\end{table}

\subsubsection{Necessity of Multi-Stage Distillation}
To demonstrate the necessity of multi-stage distillation, we conduct an experiment named E2 by enlarging training iterations of stage 1 to be equal to proposed three-stage learning procedure, where stage 1 is optimized with total 700k iterations for fair comparison. Compare E2 and E1 in Table~\ref{tab:4}, flow accuracy on diverse datasets has almost no improvement. We attribute the reason to be that error of pseudo label in augmentation regularization term has impeded the optimization of Eq.~\ref{eq:4} to trap into local minimum. Therefore, multi-stage distilation is necessary.

\subsubsection{One More Mutual Knowledge Distillation}
Since proposed approach can bring progressive improvement between the teacher and the student, we carry out an experiment to perform one more mutual distilation between these two networks, whose forward and backward distillation stages are denoted as E12 and E13 in Table~\ref{tab:4}, respectively. As is expected, compare E12 with E8, and compare E13 with E11, we can see that both the student and the teacher have achieved better flow accuracy against their previous optimization stage. However, the reduction of average endpoint error on both Sintel and KITTI are almost less than 0.1, while it requires double training iterations. Therefore, it shows that the three-stage MDFlow has already achieved relatively good results and been saturated.

\begin{table*}[t]
	\caption{\textbf{Generalization ability comparison of methods with pyramid flow architectures across different datasets.}}
	\label{tab:5}
	\centering
	\renewcommand{\arraystretch}{1.1}
	\tabcolsep=8.6mm
	\begin{tabular}{llcccccccc}
		\toprule
		\multirow{2}{*}{} & \multirow{2}{*}{Method} & C-test & \multicolumn{2}{c}{S-train (EPE)} & \multicolumn{2}{c}{K-15-train} \\
		\cmidrule(lr){3-3}
		\cmidrule(lr){4-5}
		\cmidrule(lr){6-7}
		& & EPE & Clean & Final & EPE & Fl-all \\
		\midrule
		\parbox[t]{2mm}{\multirow{4}{*}{\rotatebox{90}{Chairs}}}
		& DDFlow~\cite{Liu:2019:DDFlow} &2.97 & 4.83 & 4.85 & 17.26 & - \\
		& SimFlow~\cite{10.1007/978-3-030-58586-0_11} & 2.69 & 3.66 & 4.67 & 16.99 & - \\
		& UFlow~\cite{10.1007/978-3-030-58536-5_33} & 2.55 & 3.43 & 4.17 & 11.27 & 30.31\% \\
		& MDFlow (Ours) & \textbf{2.48} & \textbf{2.89} & \textbf{4.00} & \textbf{9.60} & \textbf{25.87\%} \\
		\hline
		\parbox[t]{2mm}{\multirow{5}{*}{\rotatebox{90}{Sintel}}}
		& DDFlow~\cite{Liu:2019:DDFlow} & 3.46 & (2.92) & (3.98) & 12.69 & - \\
		& ARFlow~\cite{Liu_2020_CVPR} & 3.50 & 2.79 & 3.73 & 9.04 & - \\
		& SimFlow~\cite{10.1007/978-3-030-58586-0_11} & 3.01 & (2.86) & (3.57) & 12.75 & - \\
		& UFlow~\cite{10.1007/978-3-030-58536-5_33} & 3.25 & (2.50) & (3.39) & 9.40 & 20.02\% \\
		& MDFlow (Ours) & \textbf{2.79} & (\textbf{2.17}) & (\textbf{3.14}) & \textbf{5.92} & \textbf{16.00\%} \\
		\hline
		\parbox[t]{2mm}{\multirow{4}{*}{\rotatebox{90}{KITTI}}}
		& DDFlow~\cite{Liu:2019:DDFlow} & 6.35 & 6.20 & 7.08 & 5.72 & - \\
		& SimFlow~\cite{10.1007/978-3-030-58586-0_11} & 4.32 & 5.49 & 7.24 & 5.19 & - \\
		& UFlow~\cite{10.1007/978-3-030-58536-5_33} & 5.05 & 5.58 & 6.31 & 2.71 & 9.05\% \\
		& MDFlow (Ours) & \textbf{4.14} & \textbf{4.04} & \textbf{5.34} & \textbf{2.45} & \textbf{8.10\%} \\
		\bottomrule
	\end{tabular}
\end{table*}

In short, our proposed MDFlow (E11) improves the baseline (E1) results by $9.2\%$, $6.0\%$ and $16.0\%$ on Sintel Clean, Sintel Final and KITTI 2015 training sets respectively, that is consistent with the improvement over the comparable state-of-the-art UFlow~\cite{10.1007/978-3-030-58536-5_33} on multiple test benchmarks.

\begin{figure}[t]
	\vspace{0.7mm}
	\footnotesize
	\centering
	\resizebox{0.49\textwidth}{!}{
		\begin{tabular}{@{}c @{\hskip 0.01in} c @{\hskip 0.01in} c @{\hskip 0.01in} c @{\hskip 0.01in} c@{}}
			\includegraphics[width=0.195\linewidth]{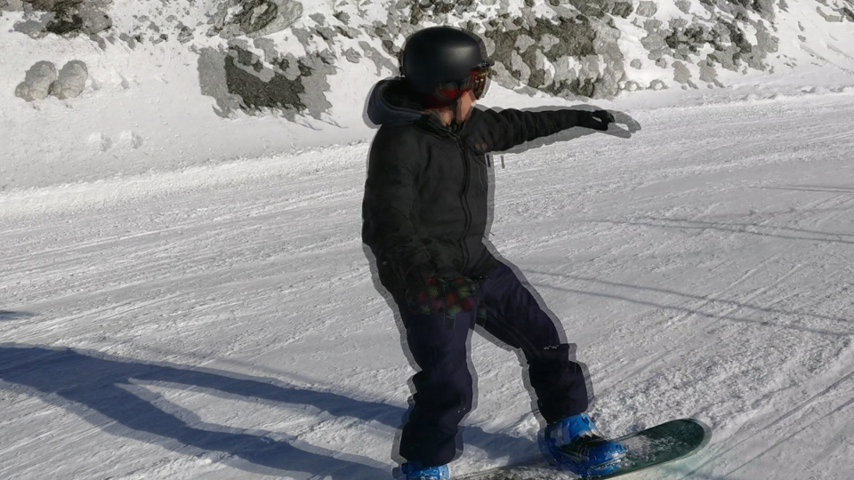}
			&
			\includegraphics[width=0.195\linewidth]{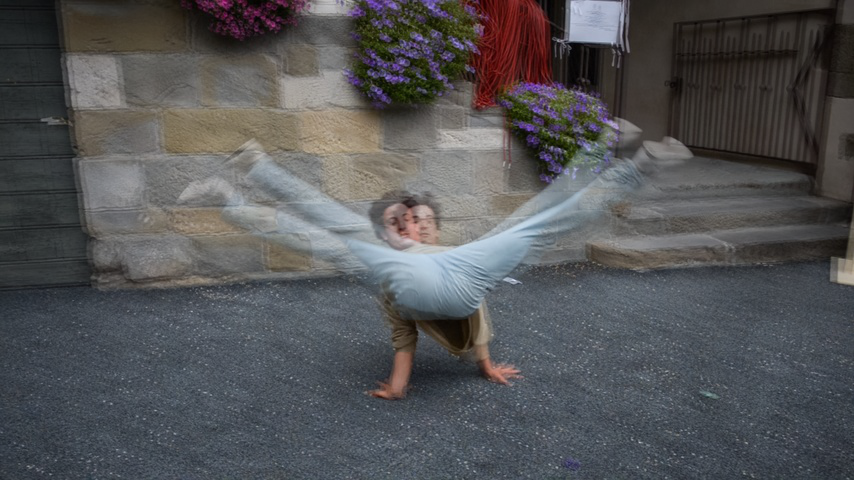}
			&
			\includegraphics[width=0.195\linewidth]{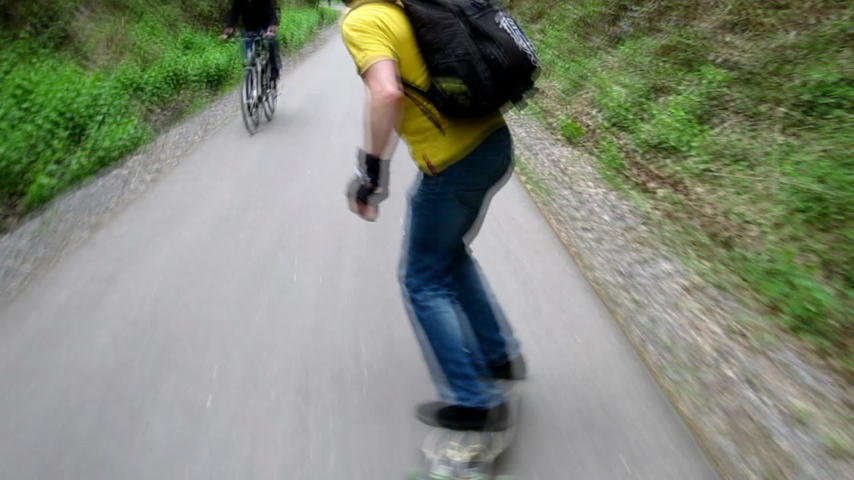}
			\vspace{-0.7mm}
			\\
			\includegraphics[width=0.195\linewidth]{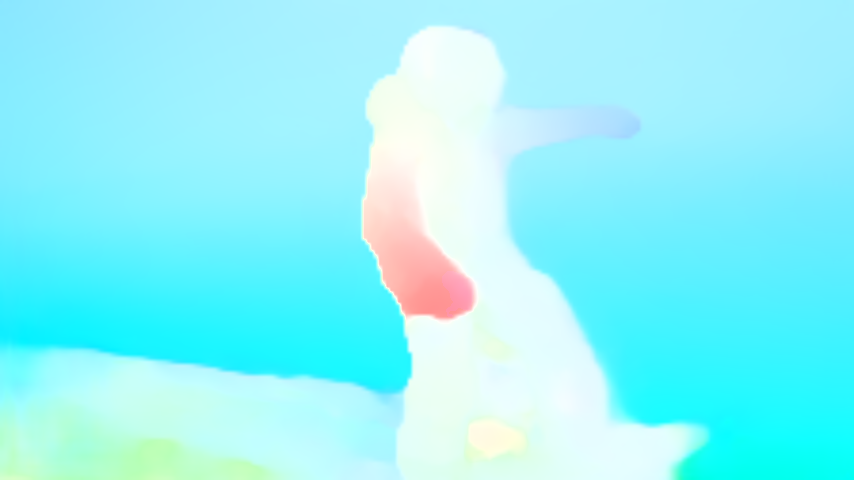}
			&
			\includegraphics[width=0.195\linewidth]{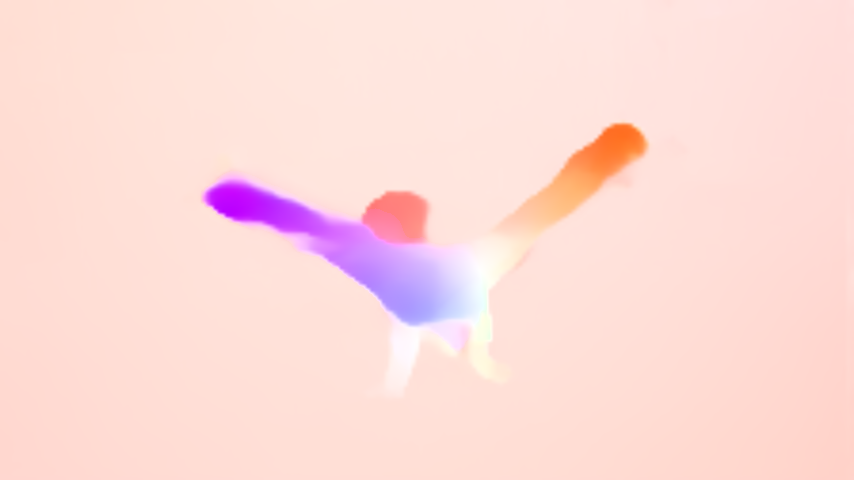}
			&
			\includegraphics[width=0.195\linewidth]{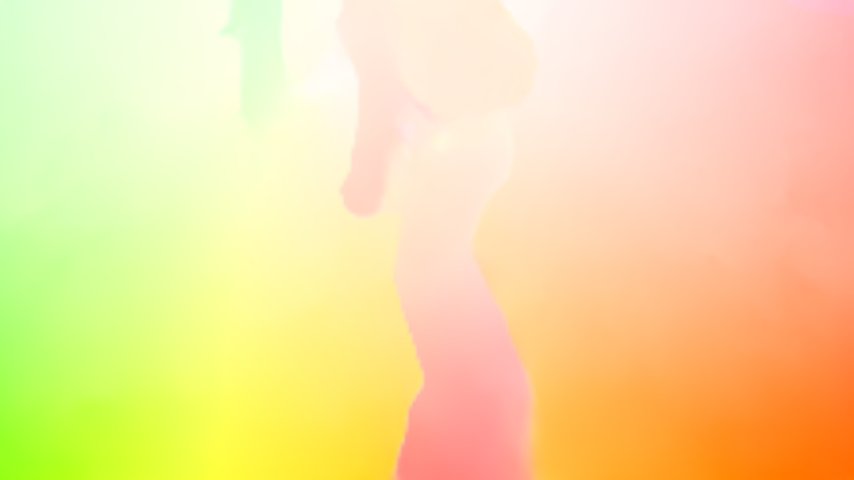}
			\vspace{-0.7mm}
			\\
	\end{tabular}}
	\caption{\textbf{Qualitative results of the teacher network on the cross domain DAVIS dataset~\cite{Jordi_arXiv_2017}.} Three pairs of image and flow examples are shown.}
	\label{fig:6}
\end{figure}

\subsection{Cross Domain Generalization}
Finally, we conduct cross domain experiments to compare with recent top-performing unsupervised methods. As illustrated in Table~\ref{tab:5}, our method consistently performs best when training on one dataset and testing on another one, which demonstrates the state-of-the-art generalization ability of MDFlow among pyramid-based approaches. We attribute its excellent generalization performance to the efficient utilization of data augmentation on reliable pseudo labels during the decoupled mutual knowledge distillation process. Figure~\ref{fig:6} presents qualitative results of the efficient teacher network on the cross domain DAVIS dataset~\cite{Jordi_arXiv_2017}. It can be seen that proposed unsupervised learning approaches can estimate relatively good optical flow with clear motion boundary in diverse dynamic scenes.

\section{Conclusion}
To our best knowledge, it is the first time that mutual knowledge distillation framework is introduced to unsupervised optical flow, which can efficiently leverage countless unlabeled video sequences for optical flow learning. To decouple mismatched pseudo labels that block the learning process, we propose a confidence matching selection mechanism to partly exclude influence from outliers. Then, we use diverse augmentation like supervised methods and above reliable pseudo labels to train a stronger student model for more accurate flow prediction. Finally, a novel multi-target backward distillation procedure is built to transfer better student knowledge back to the efficient teacher without sacrificing generalization ability. Experiments on Flying Chairs, MPI Sintel and KITTI 2015 datasets show that our framework achieves state-of-the-art real-time accuracy and generalization performance. In the future, we plan to explore proposed reliable mutual knowledge distillation approach on other unsupervised matching tasks, such as Structure from Motion (SfM) and scene flow estimation.

\bibliographystyle{IEEEtran}
\bibliography{reference}

\begin{thebibliography}{10}
\providecommand{\url}[1]{#1}
\csname url@samestyle\endcsname
\providecommand{\newblock}{\relax}
\providecommand{\bibinfo}[2]{#2}
\providecommand{\BIBentrySTDinterwordspacing}{\spaceskip=0pt\relax}
\providecommand{\BIBentryALTinterwordstretchfactor}{4}
\providecommand{\BIBentryALTinterwordspacing}{\spaceskip=\fontdimen2\font plus
\BIBentryALTinterwordstretchfactor\fontdimen3\font minus
  \fontdimen4\font\relax}
\providecommand{\BIBforeignlanguage}[2]{{%
\expandafter\ifx\csname l@#1\endcsname\relax
\typeout{** WARNING: IEEEtran.bst: No hyphenation pattern has been}%
\typeout{** loaded for the language `#1'. Using the pattern for}%
\typeout{** the default language instead.}%
\else
\language=\csname l@#1\endcsname
\fi
#2}}
\providecommand{\BIBdecl}{\relax}
\BIBdecl

\bibitem{9241798}
C.~Wang, X.~Chen, S.~Min, J.~Wang, and Z.-J. Zha, ``Structure-guided deep video
  inpainting,'' \emph{IEEE Transactions on Circuits and Systems for Video
  Technology}, 2021.

\bibitem{Kong_2022_CVPR}
L.~Kong, B.~Jiang, D.~Luo, W.~Chu, X.~Huang, Y.~Tai, C.~Wang, and J.~Yang,
  ``Ifrnet: Intermediate feature refine network for efficient frame
  interpolation,'' in \emph{Proceedings of the IEEE/CVF Conference on Computer
  Vision and Pattern Recognition (CVPR)}, 2022.

\bibitem{Liu_2022_ICIP}
J.~Liu, L.~Kong, and J.~Yang, ``Atca: an arc trajectory based model with
  curvature attention for video frame interpolation,'' in \emph{2022 IEEE
  International Conference on Image Processing (ICIP)}, 2022.

\bibitem{7508986}
J.~Dong and H.~Liu, ``Video stabilization for strict real-time applications,''
  \emph{IEEE Transactions on Circuits and Systems for Video Technology}, 2017.

\bibitem{10.5555/888857}
B.~K. Horn and B.~G. Schunck, ``Determining optical flow,'' \emph{Artificial
  Intelligence}, 1981.

\bibitem{5206697}
T.~Brox, C.~Bregler, and J.~Malik, ``Large displacement optical flow,'' in
  \emph{2009 IEEE Conference on Computer Vision and Pattern Recognition}, 2009.

\bibitem{5539939}
D.~Sun, S.~Roth, and M.~J. Black, ``Secrets of optical flow estimation and
  their principles,'' in \emph{2010 IEEE Computer Society Conference on
  Computer Vision and Pattern Recognition}, 2010.

\bibitem{8434339}
C.~Bailer, B.~Taetz, and D.~Stricker, ``Flow fields: Dense correspondence
  fields for highly accurate large displacement optical flow estimation,''
  \emph{IEEE Transactions on Pattern Analysis and Machine Intelligence}, 2019.

\bibitem{7410673}
A.~Dosovitskiy, P.~Fischer, E.~Ilg, P.~Häusser, C.~Hazirbas, V.~Golkov,
  P.~v.~d. Smagt, D.~Cremers, and T.~Brox, ``Flownet: Learning optical flow
  with convolutional networks,'' in \emph{2015 IEEE International Conference on
  Computer Vision (ICCV)}, 2015.

\bibitem{8099662}
E.~Ilg, N.~Mayer, T.~Saikia, M.~Keuper, A.~Dosovitskiy, and T.~Brox, ``Flownet
  2.0: Evolution of optical flow estimation with deep networks,'' in \emph{2017
  IEEE Conference on Computer Vision and Pattern Recognition (CVPR)}, 2017.

\bibitem{8579029}
D.~Sun, X.~Yang, M.-Y. Liu, and J.~Kautz, ``Pwc-net: Cnns for optical flow
  using pyramid, warping, and cost volume,'' in \emph{2018 IEEE/CVF Conference
  on Computer Vision and Pattern Recognition}, 2018.

\bibitem{Hur:2019:IRR}
J.~Hur and S.~Roth, ``Iterative residual refinement for joint optical flow and
  occlusion estimation,'' in \emph{CVPR}, 2019.

\bibitem{teed2020raft}
Z.~Teed and J.~Deng, ``Raft: Recurrent all-pairs field transforms for optical
  flow,'' in \emph{{European Conference on Computer Vision (ECCV)}}, 2020.

\bibitem{NEURIPS2020_add5aebf}
J.~Wang, Y.~Zhong, Y.~Dai, K.~Zhang, P.~Ji, and H.~Li, ``Displacement-invariant
  matching cost learning for accurate optical flow estimation,'' in
  \emph{Advances in Neural Information Processing Systems}, 2020.

\bibitem{7780807}
N.~Mayer, E.~Ilg, P.~Häusser, P.~Fischer, D.~Cremers, A.~Dosovitskiy, and
  T.~Brox, ``A large dataset to train convolutional networks for disparity,
  optical flow, and scene flow estimation,'' in \emph{2016 IEEE Conference on
  Computer Vision and Pattern Recognition (CVPR)}, 2016.

\bibitem{humanflow}
A.~Ranjan, J.~Romero, and M.~J. Black, ``Learning human optical flow,'' in
  \emph{29th British Machine Vision Conference}, Sep. 2018.

\bibitem{flowdata}
N.~Mayer, E.~Ilg, P.~Fischer, C.~Hazirbas, D.~Cremers, A.~Dosovitskiy, and
  T.~Brox, ``What makes good synthetic training data for learning disparity and
  optical flow estimation?'' \emph{International Journal of Computer Vision},
  2018.

\bibitem{10.1007/978-3-319-49409-8_1}
J.~J. Yu, A.~W. Harley, and K.~G. Derpanis, ``Back to basics: Unsupervised
  learning of optical flow via brightness constancy and motion smoothness,'' in
  \emph{Computer Vision -- ECCV 2016 Workshops}, 2016.

\bibitem{10.5555/3298239.3298457}
Z.~Ren, J.~Yan, B.~Ni, B.~Liu, X.~Yang, and H.~Zha, ``Unsupervised deep
  learning for optical flow estimation,'' in \emph{Proceedings of the
  Thirty-First AAAI Conference on Artificial Intelligence}, ser. AAAI'17, 2017.

\bibitem{Meister:2018:UUL}
S.~Meister, J.~Hur, and S.~Roth, ``{UnFlow}: Unsupervised learning of optical
  flow with a bidirectional census loss,'' in \emph{AAAI}, Feb. 2018.

\bibitem{8578611}
Y.~Wang, Y.~Yang, Z.~Yang, L.~Zhao, P.~Wang, and W.~Xu, ``Occlusion aware
  unsupervised learning of optical flow,'' in \emph{2018 IEEE/CVF Conference on
  Computer Vision and Pattern Recognition}, 2018.

\bibitem{Liu_2020_CVPR}
L.~Liu, J.~Zhang, R.~He, Y.~Liu, Y.~Wang, Y.~Tai, D.~Luo, C.~Wang, J.~Li, and
  F.~Huang, ``Learning by analogy: Reliable supervision from transformations
  for unsupervised optical flow estimation,'' in \emph{Proceedings of the
  IEEE/CVF Conference on Computer Vision and Pattern Recognition (CVPR)}, June
  2020.

\bibitem{10.1007/978-3-030-58536-5_33}
R.~Jonschkowski, A.~Stone, J.~T. Barron, A.~Gordon, K.~Konolige, and
  A.~Angelova, ``What matters in unsupervised optical flow,'' in \emph{Computer
  Vision -- ECCV 2020}, 2020.

\bibitem{Luo_2021_CVPR}
K.~Luo, C.~Wang, S.~Liu, H.~Fan, J.~Wang, and J.~Sun, ``Upflow: Upsampling
  pyramid for unsupervised optical flow learning,'' in \emph{Proceedings of the
  IEEE/CVF Conference on Computer Vision and Pattern Recognition (CVPR)}, June
  2021.

\bibitem{Liu:2019:SelFlow}
P.~Liu, M.~R. Lyu, I.~King, and J.~Xu, ``Selflow: Self-supervised learning of
  optical flow,'' in \emph{CVPR}, 2019.

\bibitem{10.1007/978-3-030-58586-0_11}
W.~Im, T.-K. Kim, and S.-E. Yoon, ``Unsupervised learning of optical flow with
  deep feature similarity,'' in \emph{Computer Vision -- ECCV 2020}, 2020.

\bibitem{Stone_2021_CVPR}
A.~Stone, D.~Maurer, A.~Ayvaci, A.~Angelova, and R.~Jonschkowski, ``Smurf:
  Self-teaching multi-frame unsupervised raft with full-image warping,'' in
  \emph{Proceedings of the IEEE/CVF Conference on Computer Vision and Pattern
  Recognition (CVPR)}, 2021.

\bibitem{10.1007/BFb0028345}
R.~Zabih and J.~Woodfill, ``Non-parametric local transforms for computing
  visual correspondence,'' in \emph{Computer Vision --- ECCV '94}, 1994.

\bibitem{Butler:ECCV:2012}
D.~J. Butler, J.~Wulff, G.~B. Stanley, and M.~J. Black, ``A naturalistic open
  source movie for optical flow evaluation,'' in \emph{European Conf. on
  Computer Vision (ECCV)}, Oct. 2012, pp. 611--625.

\bibitem{Menze2015CVPR}
M.~Menze and A.~Geiger, ``Object scene flow for autonomous vehicles,'' in
  \emph{Conference on Computer Vision and Pattern Recognition (CVPR)}, 2015.

\bibitem{7780984}
Y.~Hu, R.~Song, and Y.~Li, ``Efficient coarse-to-fine patch match for large
  displacement optical flow,'' in \emph{2016 IEEE Conference on Computer Vision
  and Pattern Recognition (CVPR)}, 2016.

\bibitem{8099774}
A.~Ranjan and M.~J. Black, ``Optical flow estimation using a spatial pyramid
  network,'' in \emph{2017 IEEE Conference on Computer Vision and Pattern
  Recognition (CVPR)}, 2017.

\bibitem{8579034}
T.-W. Hui, X.~Tang, and C.~C. Loy, ``Liteflownet: A lightweight convolutional
  neural network for optical flow estimation,'' in \emph{2018 IEEE/CVF
  Conference on Computer Vision and Pattern Recognition}, 2018.

\bibitem{8846749}
M.~Zhai, X.~Xiang, R.~Zhang, N.~Lv, and A.~El~Saddik, ``Optical flow estimation
  using dual self-attention pyramid networks,'' \emph{IEEE Transactions on
  Circuits and Systems for Video Technology}, 2020.

\bibitem{zhao2020maskflownet}
S.~Zhao, Y.~Sheng, Y.~Dong, E.~I.-C. Chang, and Y.~Xu, ``Maskflownet:
  Asymmetric feature matching with learnable occlusion mask,'' in
  \emph{Proceedings of the IEEE Conference on Computer Vision and Pattern
  Recognition (CVPR)}, 2020.

\bibitem{9413531}
L.~Kong, X.~Yang, and J.~Yang, ``Oas-net: Occlusion aware sampling network for
  accurate optical flow,'' in \emph{ICASSP 2021 - 2021 IEEE International
  Conference on Acoustics, Speech and Signal Processing (ICASSP)}, 2021.

\bibitem{9191101}
L.~Kong and J.~Yang, ``Fdflownet: Fast optical flow estimation using a deep
  lightweight network,'' in \emph{2020 IEEE International Conference on Image
  Processing (ICIP)}, 2020.

\bibitem{Kong_2021_ICRA}
L.~Kong, C.~Shen, and J.~Yang, ``Fastflownet: A lightweight network for fast
  optical flow estimation,'' in \emph{2021 IEEE International Conference on
  Robotics and Automation (ICRA)}, 2021.

\bibitem{Jiang_2021_ICCV}
S.~Jiang, D.~Campbell, Y.~Lu, H.~Li, and R.~Hartley, ``Learning to estimate
  hidden motions with global motion aggregation,'' in \emph{Proceedings of the
  IEEE/CVF International Conference on Computer Vision (ICCV)}, 2021.

\bibitem{Liu:2019:DDFlow}
P.~Liu, I.~King, M.~R. Lyu, and J.~Xu, ``Ddflow: Learning optical flow with
  unlabeled data distillation,'' in \emph{AAAI}, 2019.

\bibitem{9201360}
Z.~Ren, W.~Luo, J.~Yan, W.~Liao, X.~Yang, A.~Yuille, and H.~Zha, ``Stflow:
  Self-taught optical flow estimation using pseudo labels,'' \emph{IEEE
  Transactions on Image Processing}, 2020.

\bibitem{Wang_CoRL_2020}
H.~Wang, R.~Fan, and M.~Liu, ``Cot-amflow: Adaptive modulation network with
  co-teaching strategy for unsupervised optical flow estimation,'' in \emph{4th
  Conference on Robot Learning, CoRL 2020, 16-18 November 2020, Virtual Event /
  Cambridge, MA, USA}, 2020.

\bibitem{9444870}
P.~Liu, M.~R. Lyu, I.~King, and J.~Xu, ``Learning by distillation: A
  self-supervised learning framework for optical flow estimation,'' \emph{IEEE
  Transactions on Pattern Analysis and Machine Intelligence}, 2021.

\bibitem{9477059}
S.~Liu, K.~Luo, N.~Ye, C.~Wang, J.~Wang, and B.~Zeng, ``Oiflow:
  Occlusion-inpainting optical flow estimation by unsupervised learning,''
  \emph{IEEE Transactions on Image Processing}, 2021.

\bibitem{9625946}
S.~Liu, K.~Luo, A.~Luo, C.~Wang, F.~Meng, and B.~Zeng, ``Asflow: Unsupervised
  optical flow learning with adaptive pyramid sampling,'' \emph{IEEE
  Transactions on Circuits and Systems for Video Technology}, 2021.

\bibitem{9648363}
R.~Zhao, R.~Xiong, Z.~Ding, X.~Fan, J.~Zhang, and T.~Huang, ``Mrdflow:
  Unsupervised optical flow estimation network with multi-scale recurrent
  decoder,'' \emph{IEEE Transactions on Circuits and Systems for Video
  Technology}, 2021.

\bibitem{Wang_2019_CVPR}
Y.~Wang, P.~Wang, Z.~Yang, C.~Luo, Y.~Yang, and W.~Xu, ``Unos: Unified
  unsupervised optical-flow and stereo-depth estimation by watching videos,''
  in \emph{Proceedings of the IEEE/CVF Conference on Computer Vision and
  Pattern Recognition (CVPR)}, 2019.

\bibitem{Flow2Stereo}
P.~Liu, I.~King, M.~R. Lyu, and J.~Xu, ``Flow2stereo: Effective self-supervised
  learning of optical flow and stereo matching,'' in \emph{Proceedings of the
  IEEE/CVF Conference on Computer Vision and Pattern Recognition (CVPR)}, 2020.

\bibitem{Chi_2021_CVPR}
C.~Chi, Q.~Wang, T.~Hao, P.~Guo, and X.~Yang, ``Feature-level collaboration:
  Joint unsupervised learning of optical flow, stereo depth and camera
  motion,'' in \emph{Proceedings of the IEEE/CVF Conference on Computer Vision
  and Pattern Recognition (CVPR)}, 2021.

\bibitem{Li_2021_ICCV}
H.~Li, K.~Luo, and S.~Liu, ``Gyroflow: Gyroscope-guided unsupervised optical
  flow learning,'' in \emph{Proceedings of the IEEE/CVF International
  Conference on Computer Vision (ICCV)}, 2021.

\bibitem{44873}
G.~Hinton, O.~Vinyals, and J.~Dean, ``Distilling the knowledge in a neural
  network,'' in \emph{NIPS Deep Learning and Representation Learning Workshop},
  2015.

\bibitem{DBLP:journals/corr/RomeroBKCGB14}
A.~Romero, N.~Ballas, S.~E. Kahou, A.~Chassang, C.~Gatta, and Y.~Bengio,
  ``Fitnets: Hints for thin deep nets,'' in \emph{3rd International Conference
  on Learning Representations, {ICLR} 2015, San Diego, CA, USA, May 7-9, 2015,
  Conference Track Proceedings}, 2015.

\bibitem{Zagoruyko2017AT}
S.~Zagoruyko and N.~Komodakis, ``Paying more attention to attention: Improving
  the performance of convolutional neural networks via attention transfer,'' in
  \emph{ICLR}, 2017.

\bibitem{8953814}
D.~Sun, A.~Yao, A.~Zhou, and H.~Zhao, ``Deeply-supervised knowledge synergy,''
  in \emph{2019 IEEE/CVF Conference on Computer Vision and Pattern Recognition
  (CVPR)}, 2019.

\bibitem{tian2019crd}
Y.~Tian, D.~Krishnan, and P.~Isola, ``Contrastive representation
  distillation,'' in \emph{International Conference on Learning
  Representations}, 2020.

\bibitem{Zhang_2018_CVPR}
Y.~Zhang, T.~Xiang, T.~M. Hospedales, and H.~Lu, ``Deep mutual learning,'' in
  \emph{Proceedings of the IEEE Conference on Computer Vision and Pattern
  Recognition (CVPR)}, 2018.

\bibitem{9461003}
K.~Zhang, C.~Zhang, S.~Li, D.~Zeng, and S.~Ge, ``Student network learning via
  evolutionary knowledge distillation,'' \emph{IEEE Transactions on Circuits
  and Systems for Video Technology}, 2022.

\bibitem{10.1007/978-3-030-58555-6_18}
A.~Yao and D.~Sun, ``Knowledge transfer via dense cross-layer
  mutual-distillation,'' in \emph{Computer Vision -- ECCV 2020}, 2020.

\bibitem{NEURIPS2021_c203d8a1}
B.~Zhao and K.~Han, ``Novel visual category discovery with dual ranking
  statistics and mutual knowledge distillation,'' in \emph{Advances in Neural
  Information Processing Systems}.\hskip 1em plus 0.5em minus 0.4em\relax
  Curran Associates, Inc., 2021.

\bibitem{8954274}
R.~Wu, M.~Feng, W.~Guan, D.~Wang, H.~Lu, and E.~Ding, ``A mutual learning
  method for salient object detection with intertwined multi-supervision,'' in
  \emph{2019 IEEE/CVF Conference on Computer Vision and Pattern Recognition
  (CVPR)}, 2019.

\bibitem{9577564}
Q.~Zhai, X.~Li, F.~Yang, C.~Chen, H.~Cheng, and D.-P. Fan, ``Mutual graph
  learning for camouflaged object detection,'' in \emph{2021 IEEE/CVF
  Conference on Computer Vision and Pattern Recognition (CVPR)}, 2021.

\bibitem{9710140}
A.~Hao, Y.~Min, and X.~Chen, ``Self-mutual distillation learning for continuous
  sign language recognition,'' in \emph{2021 IEEE/CVF International Conference
  on Computer Vision (ICCV)}, 2021.

\bibitem{9782430}
B.~Hu, S.~Zhou, Z.~Xiong, and F.~Wu, ``Cross-resolution distillation for
  efficient 3d medical image registration,'' \emph{IEEE Transactions on
  Circuits and Systems for Video Technology}, 2022.

\bibitem{10.1007/978-3-642-15549-9_32}
N.~Sundaram, T.~Brox, and K.~Keutzer, ``Dense point trajectories by
  gpu-accelerated large displacement optical flow,'' in \emph{Computer Vision
  -- ECCV 2010}, 2010.

\bibitem{10.1007/978-3-030-01234-2_40}
E.~Ilg, {\"O}.~{\c{C}}i{\c{c}}ek, S.~Galesso, A.~Klein, O.~Makansi, F.~Hutter,
  and T.~Brox, ``Uncertainty estimates and multi-hypotheses networks for
  optical flow,'' in \emph{Computer Vision -- ECCV 2018}, 2018.

\bibitem{aleotti2020learning}
F.~Aleotti, M.~Poggi, F.~Tosi, and S.~Mattoccia, ``Learning end-to-end scene
  flow by distilling single tasks knowledge,'' in \emph{Thirty-Fourth AAAI
  Conference on Artificial Intelligence}, 2020.

\bibitem{8621052}
D.~Sun, X.~Yang, M.-Y. Liu, and J.~Kautz, ``Models matter, so does training: An
  empirical study of cnns for optical flow estimation,'' \emph{IEEE
  Transactions on Pattern Analysis and Machine Intelligence}, 2020.

\bibitem{Bar-Haim_2020_CVPR}
A.~Bar-Haim and L.~Wolf, ``Scopeflow: Dynamic scene scoping for optical flow,''
  in \emph{The IEEE/CVF Conference on Computer Vision and Pattern Recognition
  (CVPR)}, June 2020.

\bibitem{kingma2014method}
D.~P. Kingma and J.~Ba, ``Adam: A method for stochastic optimization,'' in
  \emph{The 3rd International Conference on Learning Representations, San
  Diego, 2015}, 2015.

\bibitem{loshchilov2018decoupled}
I.~Loshchilov and F.~Hutter, ``Decoupled weight decay regularization,'' in
  \emph{International Conference on Learning Representations}, 2019.

\bibitem{Janai_2018_ECCV}
J.~Janai, F.~Guney, A.~Ranjan, M.~Black, and A.~Geiger, ``Unsupervised learning
  of multi-frame optical flow with occlusions,'' in \emph{Proceedings of the
  European Conference on Computer Vision (ECCV)}, September 2018.

\bibitem{8953885}
Y.~Zhong, P.~Ji, J.~Wang, Y.~Dai, and H.~Li, ``Unsupervised deep epipolar flow
  for stationary or dynamic scenes,'' in \emph{2019 IEEE/CVF Conference on
  Computer Vision and Pattern Recognition (CVPR)}, 2019.

\bibitem{Jordi_arXiv_2017}
J.~Pont-Tuset, F.~Perazzi, S.~Caelles, P.~Arbel\'aez, A.~Sorkine-Hornung, and
  L.~V. Gool, ``The 2017 davis challenge on video object segmentation,''
  \emph{arXiv:1704.00675}, 2017.

\end{thebibliography}

\vfill

\end{document}